\documentclass[sigconf]{acmart}

\usepackage{algorithm}
\usepackage{algorithmic}

\usepackage{subfigure}
\usepackage{amsmath}
\usepackage{amsfonts}
\usepackage{booktabs}
\usepackage{enumitem}

\usepackage{graphicx} 
\usepackage{multirow} 
\usepackage{booktabs}
\usepackage{amsmath}
\usepackage{makecell}
\usepackage{caption}
\usepackage{amssymb}
\newtheorem{Proposition}{Proposition}
\usepackage{subfigure}
\usepackage{graphicx}
\usepackage{algorithm}
\usepackage{algorithmic}
\usepackage{bbding}

\newcommand{\wrt}{\textit{w.r.t. }}

\newcommand{\aka}{\textit{aka. }}

\AtBeginDocument{%
  }

\setcopyright{acmlicensed}
\copyrightyear{2018}
\acmYear{2018}
\acmDOI{XXXXXXX.XXXXXXX}
\acmConference[Conference acronym 'XX]{Make sure to enter the correct
  conference title from your rights confirmation emai}{June 03--05,
  2018}{Woodstock, NY}
\acmISBN{978-1-4503-XXXX-X/18/06}

\begin{document}

\title{Addressing Graph Heterogeneity and Heterophily from A Spectral Perspective}


\author{Kangkang Lu}
\affiliation{%
  \institution{Beijing University of Posts and Telecommunications}
  \city{Beijing}
  \country{China}}
\email{lukangkang@bupt.edu.cn}

\author{Yanhua Yu}
\affiliation{%
  \institution{Beijing University of Posts and Telecommunications}
  \city{Beijing}
  \country{China}}
\email{yuyanhua@bupt.edu.cn}

\author{Zhiyong Huang}
\affiliation{%
  \institution{National University of Singapore}
  \city{Singapore}
  \country{Singapore}}
\email{huangzy@comp.nus.edu.sg}

\author{Yunshan Ma}
\affiliation{%
  \institution{National University of Singapore}
  \city{Singapore}
  \country{Singapore}}
\email{yunshan.ma@u.nus.edu}

\author{Xiao Wang}
\affiliation{%
  \institution{Beihang University}
  \city{Beijing}
  \country{China}}
\email{xiao_wang@buaa.edu.cn}

\author{Meiyu Liang}
\affiliation{%
  \institution{Beijing University of Posts and Telecommunications}
  \city{Beijing}
  \country{China}}
\email{meiyu1210@bupt.edu.cn}

\author{Yuling Wang}
\affiliation{%
  \institution{Hangzhou Dianzi University}
  \city{Hangzhou}
  \country{China}}
\email{wangyl0612@hdu.edu.cn}

\author{Yimeng Ren}
\affiliation{%
  \institution{Beijing University of Posts and Telecommunications}
  \city{Beijing}
  \country{China}}
\email{renyimeng@bupt.edu.cn}

\author{Tat-Seng Chua}
\affiliation{%
  \institution{National University of Singapore}
  \city{Singapore}
  \country{Singapore}}
\email{dcscts@nus.edu.sg}

\renewcommand{\shortauthors}{Trovato et al.}





\begin{abstract} 
Graph neural networks (GNNs) have demonstrated excellent performance in semi-supervised node classification tasks. Despite this, two primary challenges persist: heterogeneity and heterophily. Each of these two challenges can significantly hinder the performance of GNNs. Heterogeneity refers to a graph with multiple types of nodes or edges, while heterophily refers to the fact that connected nodes are more likely to have dissimilar attributes or labels. Although there have been few works studying heterogeneous heterophilic graphs, they either only consider the heterophily of specific meta-paths and lack expressiveness, or have high expressiveness but fail to exploit high-order neighbors. In this paper, we propose a \textbf{H}eterogeneous \textbf{H}eterophilic \textbf{S}pectral \textbf{G}raph \textbf{N}eural \textbf{N}etwork (H$^2$SGNN), which employs two modules: local independent filtering and global hybrid filtering. Local independent filtering adaptively learns node representations under different homophily, while global hybrid filtering exploits high-order neighbors to learn more possible meta-paths. Extensive experiments are conducted on four datasets to validate the effectiveness of the proposed H$^2$SGNN, which achieves superior performance with fewer parameters and memory consumption. The code is available at the GitHub repo: \url{https://github.com/Lukangkang123/H2SGNN/}.

\end{abstract}

\maketitle

\section{Introduction}
\begin{figure}[t]   
	\centering 
	\includegraphics[width=\linewidth]{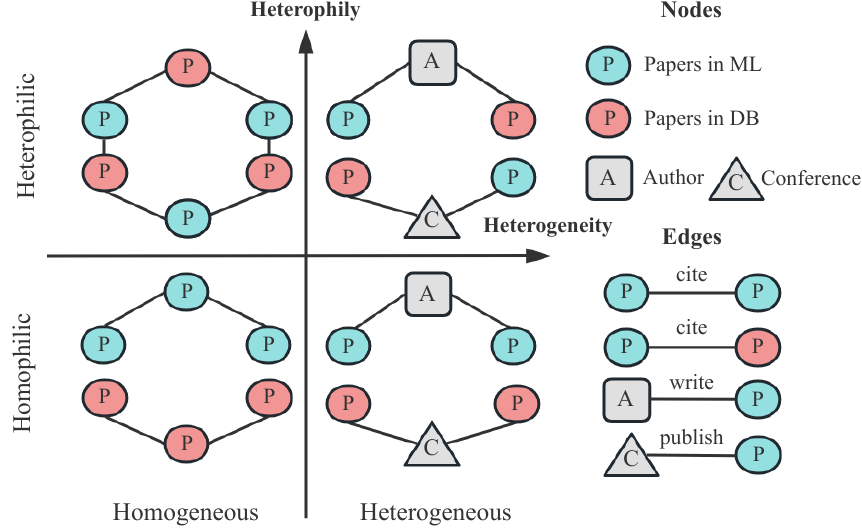}
	\caption{Example of four types of graphs based on heterogeneity and heterophily. The horizontal axis is heterogeneity, which is divided into homogeneous and heterogeneous graphs from left to right. The vertical axis is heterophily, which is divided into homophilic and heterophilic graphs from bottom to top. "Paper" is the target node to be classified.}
  \label{fig:demo}
\end{figure}

The past decade has witnessed great success and prosperity of graph neural networks (GNNs) in diverse data science and engineering scenarios, such as traffic network \cite{traffic_network_1,traffic_network_2}, abnormal detection \cite{BWGNN,tang2024gadbench}, relational databases \cite{relational_databases_1,relational_databases_2} and recommender systems \cite{graph_recommendation1,graph_recommendation2}. 
Throughout the technical development, GNNs have been confronted with two persistent challenges: heterogeneity and heterophily. Each of these two challenges can significantly impede the performance of GNNs \cite{H2GB}. Specifically, \textbf{heterogeneity} indicates multiple types of nodes or edges in a graph, which is called a \textit{heterogeneous} graph, while \textit{homogeneous} graph only has one type of node and edge. In contrast, \textbf{heterophily} characterizes a graph that how likely its connected nodes are different from each other \wrt the nodes' attributes or labels, where \textit{heterophilic} describes higher heterophily (\aka connected nodes have different labels) while \textit{homophilc} indicates lower heterophily (\aka connected nodes have similar labels). Based on these two characterizing perspectives of heterogeneity and heterophily, all the graphs can be sorted into four types: 1) homogeneous homophilic graph, 2) homogeneous heterophilic graph, 3) heterogeneous homophilic graph, and 4) heterogeneous heterophilic graph. As an example scenario of a paper citation graph where the target nodes are papers, and the edges are the citation, writing, and publication relationships, we illustrate these four types of graphs in Figure~\ref{fig:demo}. 

Among these four types of graphs, heterogeneous heterophilic graph exhibits the highest challenge while receiving the least research attention, despite their ubiquity in the real world \cite{H2GB}. Concretely, most existing works focus either on heterophily in homogeneous graphs while overlooking the diversity of node and relation types~\cite{FAGCN,GPR-GNN,Mixhop,Geom-gcn}, or on heterogeneous graphs based on the homophily assumption but neglecting heterophily~\cite{HAN,magnn,RGCN,mhgcn}. On the contrary, only a few attempts have been made to study the heterogeneous heterophilic graph. Specifically, some studies \cite{HDHGR,Hetero2Net} have observed performance degradation when handling heterogeneous graphs with heterophily. To address this issue, methods such as graph rewriting \cite{HDHGR} and masked label prediction \cite{Hetero2Net} have been proposed for heterogeneous heterophilic graphs. Although they alleviate heterophily on heterogeneous graphs and bring some performance gains, they lead to suboptimal performance because they only consider a few specified meta-paths. Recently, PSHGCN \cite{PSHGCN} proposes multivariate noncommutative polynomials as filters for spectral GNNs \cite{spectral_survey,benchmarking_Spectral} to handle heterogeneous graphs. Since the adopted multivariate polynomials incorporate all possible products of relational matrices, they exhibit strong expressive power and demonstrate competitive performance. Nevertheless, PSHGCN faces the following two limitations:

\textbf{Limitation 1: Lack of explicit consideration of homophily in different meta-paths. } PSHGCN simply multiplies all relation matrices in order without accounting for the homophily specific to each meta-path\footnote{A meta-path is a predefined sequence of node and edge types. For more details, please refer to subsection \ref{sec:Heterogeneous_Graph}.}. However, in a heterogeneous graph, different meta-path subgraphs may exhibit varying degrees of homophily. For example, as shown in Table 1 for the ACM dataset, the $PAP$ (Paper - Author - Paper) and $PCP$ (Paper - Conference - Paper) meta-paths form homophilic graphs, whereas the $PKP$ (Paper - Keyword - Paper) meta-path forms a heterophilic graph. To enhance performance, meta-path subgraphs with varying homophily levels should be processed differently.


\textbf{Limitation 2: Exponential growth in the number of terms limits the utilization of higher-order neighbors.} Although a high-capacity multivariate polynomial can approximate arbitrary filters, the number of terms in a multivariate polynomial grows exponentially with the order, leading to an exponential increase in memory and parameter consumption. This makes it difficult to leverage higher-order neighbors due to high resource consumption. However, as pointed out in previous literature \cite{LMSPS, HGAMLP, gtn}, leveraging higher-order neighbors to learn more meta-paths is necessary to improve model performance in heterogeneous graphs.

To address these limitations, we propose a heterogeneous heterophilic spectral graph neural network, short as H$^2$SGNN.
First, for limitation 1, we introduce \textbf{local independent filtering}, which segments the heterogeneous graph into subgraphs based on different meta-paths. Then we perform local independent polynomial filtering on each subgraph. Thus, we can adaptively learn node representations catering to diverse degrees of homophily along with different meta-paths. Furthermore, independent polynomial filtering for each meta-path is beneficial to enhance interpretability, for example, the learned low-pass filter represents a homophilic meta-path subgraph. Second, for limitation 2, we propose \textbf{global hybrid filtering}. It measures and aggregates multiple meta-path matrices into a global adjacency matrix, on which we execute polynomial filtering operations. Theoretical analysis shows that global hybrid filtering can be equivalent to a multivariate polynomial under certain conditions. Therefore, high-order neighbors can be used to learn more meta-paths with lower resource consumption. \textbf{Our contributions are summarized as follows:}



\begin{itemize}[leftmargin=0.5cm, itemindent=0cm]
    \item We propose H$^2$SGNN, a novel spectral GNN model tailored to address heterophily and heterogeneity. H$^2$SGNN integrates local independent filtering and global hybrid filtering to improve performance comprehensively.
    \item The local independent filtering module can adaptively learn node representations of diverse homophily of different subgraphs, which enhances the interpretability of spectral GNNs on heterogeneous graphs. For instance, the learned low-pass filter represents a homophilic subgraph.
    \item The global hybrid filtering module is proven to be equivalent to a multivariate polynomial with high expressive power under certain conditions. This approach reduces the number of terms, which would exponentially increase with order, to a linear scale while maintaining expressive power. Consequently, the proposed H$^2$SGNN can enhance performance by leveraging higher-order neighbors to learn more meta-paths.


    
    
    \item We conduct extensive experiments on four datasets to validate the effectiveness of the proposed H$^2$SGNN, which achieves superior performance with fewer parameters and memory consumption.
\end{itemize}

\begin{table*}[t]
 \centering
 \caption{The homophily of different meta-path subgraphs on four datasets.}
\begin{tabular}{cccc|ccc|ccc|cc}
\toprule
dataset   & \multicolumn{3}{c|}{ACM}           & \multicolumn{3}{c|}{DBLP} & \multicolumn{3}{c|}{IMDB} & \multicolumn{2}{c}{AMiner}\\
\midrule
meta-path &  PAP   & PCP   & PKP   & APA    & APTPA  & APVPA  & MDM    & MAM    & MKM    & PAP         & PRP        \\
homophily (\%) & 81.45 & 64.03 & 33.38 & 87.22 & 32.49 & 67.00 & 40.44 & 17.26 & 13.39 & 97.16   & 86.80 \\
\bottomrule
\label{tab:homophily}
\end{tabular}
\end{table*}



\section{Preliminaries}
In this section, we first introduce spectral GNNs and homophily, and then introduce heterogeneous graphs and multivariate non-commutative polynomials defined on multiple nodes and types.
\subsection{Spectral GNN}

Assume that we have a undirected homogeneous graph $\mathcal{G}=(\mathcal{V}, \mathcal{E}, \mathbf{X})$, where $\mathcal{V}=\left\{v_1, \ldots, v_n\right\}$ denotes the vertex set of $n$ nodes, $\mathcal{E}$ is the edge set, and $\mathbf{X} \in \mathbb{R}^{n \times d}$ is node feature matrix. The corresponding adjacency matrix is $\mathbf{A} \in \{0,1\}^{n \times n}$, where $\mathbf{A}_{ij}=1$ if there is an edge between nodes $v_i$ and $v_j$, and $\mathbf{A}_{ij} =0$ otherwise. The degree matrix $\mathbf{D} = diag(d_1,...,d_n)$ of $\mathbf{A}$ is a diagonal matrix with its $i$-th diagonal entry as $d_i=\sum_j \mathbf{A}_{i j}$. The normalized Laplacian matrix $\mathbf{\hat L}=\mathbf{I} - \mathbf{D}^{-\frac{1}{2}}\mathbf{A}\mathbf{D}^{-\frac{1}{2}}$ where $\mathbf{I}$ denote the identity matrix. The normalized adjacency matrix is $\mathbf{\hat A}= \mathbf{D}^{-\frac{1}{2}}\mathbf{A}\mathbf{D}^{-\frac{1}{2}}$. Let $\mathbf{\hat L} = \mathbf{U} \boldsymbol{\Lambda} \mathbf{U}^{\top}$ denote the eigen-decomposition of $\mathbf{\hat L}$, where $\mathbf{U}$ is the matrix of eigenvectors and $\boldsymbol{\Lambda} = \text{diag} ( [\lambda_{1},\lambda_{2},\dots,\lambda_{n}])$ is the diagonal matrix of eigenvalues.

Spectral GNN is based on Fourier transform in signal processing. The Fourier transform of a graph signal \(\mathbf{x}\) is given by \(\hat{\mathbf{x}} = \mathbf{U}^{\top} \mathbf{x}\), and its inverse is expressed as \(\mathbf{x} = \mathbf{U} \hat{\mathbf{x}}\). Consequently, the graph propagation for the signal \(\mathbf{x}\) with kernel \(\mathbf{g}\) can be defined as follows:

\begin{equation}
\label{eq:Fourier_transform}
\mathbf{z}=\mathbf{g} *_{\mathcal{G}} \mathbf{x}=\mathbf{U}\left(\left(\mathbf{U}^{\top} \mathbf{g}\right) \odot \mathbf{U}^{\top} \mathbf{x}\right)=\mathbf{U} \mathbf{\hat G} \mathbf{U}^{\top} \mathbf{x},
\end{equation}
where $\mathbf{\hat G}=\operatorname{diag}\left(\hat g_1, \ldots,\hat  g_n\right)$ denotes the spectral kernel coefficients. To avoid eigen-decomposition, current works often approximate different kernels $\mathbf{H}$ using polynomial functions $h(\cdot)$ as follows:



\begin{equation}
\mathbf{H}=h(\mathbf{\hat L})= h_0 \mathbf{\hat L}^0+h_1 \mathbf{\hat L}^1+h_2 \mathbf{\hat L}^2+\cdots+h_K \mathbf{\hat L}^K
\label{eq:h_polynomial}
\end{equation}
where $K$ is the order of the polynomial $h(\cdot)$ and $h_K$ is the coefficient of the $k$-th order. Thus, Eq. (\ref{eq:Fourier_transform}) can be written as:

\begin{equation}
\begin{aligned}
\label{eq:filter}
\mathbf{Z}  & = \mathbf{H} \mathbf{X} = h(\mathbf{\hat L}) \mathbf{X}  =\mathbf{U} h(\boldsymbol{\Lambda}) \mathbf{U}^{\top}  \mathbf{X} \\
\end{aligned}
\end{equation}
where $\mathbf{Z} $ is the prediction matrix. According to Eq. (\ref{eq:h_polynomial}), Eq. (\ref{eq:filter}) and recent research \cite{spatial_spectral}, the $K$-th order polynomial in spectral GNNs is equivalent to aggregating $K$-th order neighbors in spatial GNN.

\subsection{Homophily}


The homophily metric measures the degree of association between connected nodes.  The widely adopted edge homophily \cite{edge_homo} is defined as follow:
\begin{equation}
\label{eq:homophily_edge}	
\mathcal{H}_{\text {edge }}(\mathcal{G})=\frac{1}{|\mathcal{E}|} \sum_{(u, v) \in \mathcal{E}} \mathbf{1} \left(y_u=y_v\right),
\end{equation}
where $\mathbf{1} (\cdot)$ is the indicator function, i.e., $\mathbf{1} (\cdot)=1$ if the condition holds, otherwise $\mathbf{1} (\cdot)=0$. $y_u$ is the label of node $u$, and $y_v$ is the label of node $v$ . $|\mathcal{E}|$ is the size of the edge set. Table \ref{tab:homophily} shows the edge homophily of different meta-paths on four real-world datasets.

\subsection{Heterogeneous Graph}
\label{sec:Heterogeneous_Graph}
A heterogeneous graph is defined as $\mathcal{G}=(\mathcal{V}, \mathcal{E}, \phi, \psi)$ where $\mathcal{V}$ is the set of nodes and $\mathcal{E}$ is the set of edges.  $\phi: \mathcal{V} \rightarrow \mathcal{T}_v$ maps nodes to their corresponding types, where $\mathcal{T}_v=\{\phi(v): v \in \mathcal{V}\}$. Similarly, $\psi: \mathcal{E} \rightarrow \mathcal{T}_e$ maps each edge to the type set, where $\mathcal{T}_e=\{\psi(e): e \in \mathcal{E}\}$. Specially, the graph becomes a homogeneous graph when $\lvert \mathcal{T}_v \rvert=\lvert \mathcal{T}_e \rvert=1 $.

\textbf{Metapath}. A meta-path $\mathcal{P}$ of length $n$ is defined as a sequence in the form of $A_1 \stackrel{R_1}{\longrightarrow} A_2 \stackrel{R_2}{\longrightarrow} \cdots \stackrel{R_n}{\longrightarrow} A_{n+1}$ (abbreviated as $A_1A_2 \cdots A_{n+1}$), where $A_i \in \mathcal{T}_v$ and $R_i \in \mathcal{T}_e$, describing a composite relation between node types $A_1$ and $A_{n+1}$. Especially, when $A_1=A_{n+1}$, we have an induced homogeneous subgraph $\mathcal{G}_{\mathcal{P}}$. For example, a meta-path of ``Author - Paper - Author''(APA) indicates the co-author relationship. Let \(\mathbf{A}_r\) be the adjacency matrix of the \(r\)-th type, where \(\mathbf{A}_r[i,j]\) is non-zero if there exists an edge of the \(r\)-th type from node \(i\) to node \(j\). The adjacency matrix of a meta-path is defined as the multiplication of multiple type matrices, such as: $\mathbf{A}_{APA} =  \mathbf{A}_{AP} \cdot \mathbf{A}_{PA}$.



\subsection{Multivariate Non-commutative Polynomials}


The multivariate noncommutative polynomial \cite{PSHGCN,MGNN} is the sum of the products of multiple relation matrices in order. For instance, a second-order polynomial \( h \) with two variables can be expressed as:
\begin{equation}
\begin{aligned}
h(\mathbf{A}_1, \mathbf{A}_2)
&= w_0 \mathbf{I} + w_1 \mathbf{A}_1 + w_2 \mathbf{A}_2\\
&+ w_{1,1} \mathbf{A}_1 \mathbf{A}_1 + w_{1,2} \mathbf{A}_1 \mathbf{A}_2 \\
&+ w_{2,1} \mathbf{A}_2 \mathbf{A}_1 + w_{2,2} \mathbf{A}_2 \mathbf{A}_2,
\end{aligned}
\label{eq:multivariate}
\end{equation}
where $w$ are learnable parameters, $\mathbf{A}_1$ and $\mathbf{A}_2$ are different relationship matrices. We assume that $K$ is the order and $R$ is the number of relationship matrices, then the number of terms in the multivariate polynomial is $1+ R + R^2 + \dots+R^K = \frac{R^{K+1}-1}{R-1}$.

Based on the above analysis, even though multivariate non-commutative polynomials have high expressive power because they are the sum of all possible relationship matrices, they are difficult to use higher-order neighbors to improve performance due to their exponentially increasing number of terms and parameters.


\section{Methodology}

\begin{figure*}[t] 
	\includegraphics[width=1\linewidth]{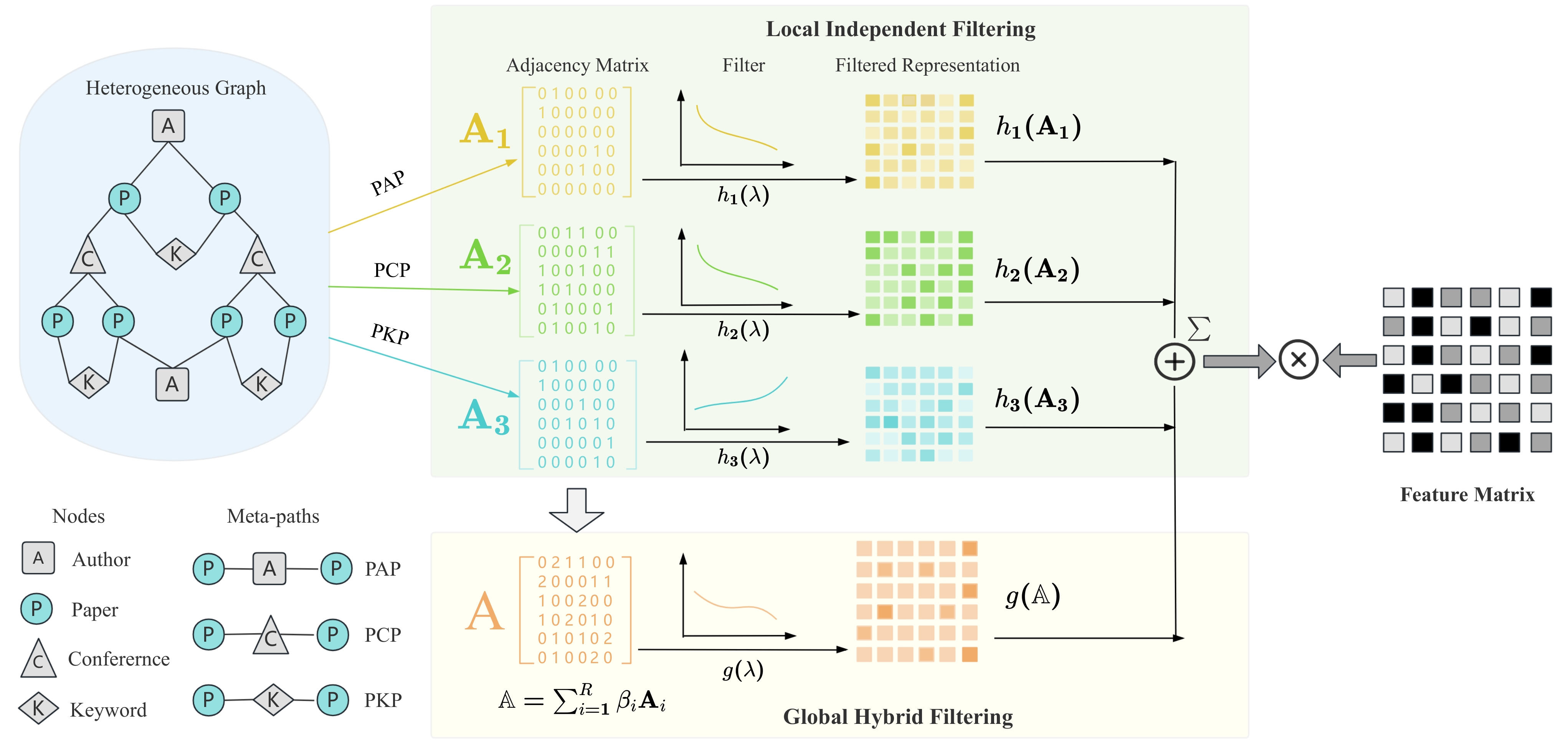}
	\caption{The overall framework of the proposed H$^2$SGNN model, where "paper" is the target node. At first, we obtain different adjacency matrices $\mathbf{A}_i$ according to different meta-paths in the heterogeneous graph, and then use different filter functions $h_i(\lambda)$ to obtain the matrix $h_i(\mathbf{A}_i)$. At the same time, the global filter function $g(\mathbb{A})$ filters the global adjacency matrix. Finally, all filtered matrices are added and multiplied with the feature matrix for the node classification task.}
	\label{fig:model}
\end{figure*}

This section describes the proposed H$^2$SGNN model. The overall architecture is shown in Figure \ref{fig:model}. In particular, the proposed H$^2$SGNN contains two key learning modules: (i) \textit{local independent filtering} and (ii) \textit{global hybrid filtering}. \textit{Local independent filtering} aims to learn node representations of meta-paths under different homophily. \textit{Global hybrid filtering} exploits high-order neighbors to learn more possible meta-paths. First, we introduce the two modules in detail. Then, we introduce the training objective and the model analysis. The overall training process and algorithm are provided in Appendix B.

\subsection{Local Individual Filtering}
\label{local_individual_filtering}

In heterogeneous graphs, the measurement of homophily is not straightforward due to different types of nodes. Therefore, we employ meta-paths of induced homogeneous subgraphs to assess homophily. Table \ref{tab:homophily} presents the edge homophily for different meta-paths on four datasets. It is evident that some meta-paths exhibit low homophily. For instance, the \textit{PKP} meta-path in the ACM dataset, the \textit{APTPA} meta-path in the DBLP dataset, and all three meta-paths in the IMDB dataset demonstrate low homophily. This indicates the necessity for more complex and diverse filters.



It is worth noting that meta-paths within the same dataset can display varying levels of homophily due to distinct connection patterns. For example, for the meta-path ``Paper - Keyword - Paper'' in the ACM dataset, two papers with the same keyword may belong to different categories, whereas for the meta-path ``Paper - Author - Paper'' in ACM dataset, two papers published by the same author often belong to the same category. As a consequence, it is imperative to apply different filters to different meta-paths. To determine the most suitable filters for each meta-path, we employ specific filter parameters for each meta-path. For the normalized adjacency matrix $\mathbf{\hat A}_{i}$ of the $i$-th meta-path subgraph, the local individual filtering operation can be expressed as follows:

\begin{equation}
\mathbf{Z}_{i}=\sum_{k=0}^K \alpha_{i,k} h_{i,k}(\mathbf{\hat A}_{i}) \mathbf{X} \mathbf{W},
\label{eq:local_individual_filtering}
\end{equation}
where $\mathbf{W} $ is a learnable weight matrix for feature mapping, $K$ is the order of the polynomial, $h_{i,k}(\cdot)$ is the $k$-th basis of the polynomial of the $i$-th meta-path, and $\alpha_{k,i}$ is the learnable coefficient corresponding to $h_{i,k}(\cdot)$. .

After filtering each meta-path separately, we add the filtered representations of each meta-path to obtain the local individual filtering representation $\mathbf{Z}_{l}$ :

\begin{equation}
\begin{aligned}
\mathbf{Z}_{l}=\sum_{i=1}^R Z_{i} = \sum_{i=1}^R \sum_{k=0}^K \alpha_{i,k} h_{i,k}(\mathbf{\hat A}_{i}) \mathbf{X} \mathbf{W},
\end{aligned}
\end{equation}
where $R$ is the number of meta-paths. In practice, we can choose different polynomial bases for the local filtering operations, such as Monomial basis GPRGNN \cite{GPR-GNN}, Jacobi basis \cite{JacobiConv} or Legendre basis \cite{Legendre}. The detailed implementation of these polynomials is shown in Appendix C.

\subsection{Global Hybrid Filtering}

\label{Global_Hybrid_Filtering}
Even though node representations can be learned through local independent filtering under different meta-paths, this approach neglects the interactions between these meta-paths and more meta-paths. For example, as shown in Figure \ref{fig:global_hybrid_filter}, local independent filtering can only obtain the results of filtering the two meta-paths separately, such as $PAPAP$ and $PCPCP$, but ignores the interaction between different meta-paths, such as $PAPCP$, $PCPAP$, etc. Consequently, after local independent filtering, it is essential to capture these interactions to learn more meta-paths. To measure the importance of each meta-path, a learnable parameter $\beta_i$ is introduced for each meta-path, facilitating the construction of a global adjacency: 
\begin{equation}
\mathbb{A}=\sum_{i=1}^{R} \beta_i \mathbf{\hat A}_i,
\label{eq:beta}
\end{equation}
where $\beta_i$ is a learnable parameter that measures the importance of meta-path $\mathbf{\hat A}_i$. Then, the representation obtained by global hybrid filtering can be expressed as:

\begin{figure}[t]   
	\centering 
	\includegraphics[width=\linewidth]{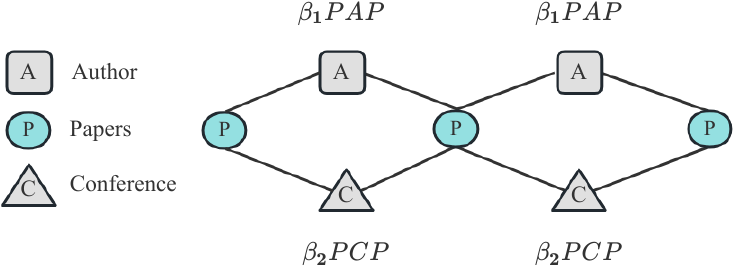}
	\caption{A simple example of a global hybrid filter exploring more diverse paths, where $\beta_1$ and $\beta_2$ are learnable parameters corresponding to the predefined meta-paths $PAP$ and $PCP$ in Eq. (\ref{eq:beta}), respectively. The learned meta-paths include $PAPAP$, $PAPCP$, $PCPAP$, and $PCPCP$ as shown in Eq. (\ref{eq:more_path}).}
  \label{fig:global_hybrid_filter}
\end{figure}

\begin{equation}
\mathbf{Z}_{g}=\sum_{k=0}^K \gamma_{k} g_k(\mathbb{A}) \mathbf{X} \mathbf{W},  
\end{equation}
where $\gamma_{k}$ is a learnable parameter and $g_k(\cdot)$ can be any polynomial basis, such as monomial basis GPRGNN \cite{GPR-GNN}, Jacobi basis \cite{JacobiConv} or Legendre basis \cite{Legendre}. Global hybrid filtering can learn more possible meta-paths. We only need to pre-define a few meta-paths, and the model can learn more possible meta-paths and automatically learn their importance without manually designing complex meta-paths. For example, as depicted in Figure \ref{fig:global_hybrid_filter}, we assume that the learnable coefficient of the matrix $\mathbf{A}_{PAP}$ formed by the meta-path $PAP$ is $\beta_1$, and the learnable coefficient of the matrix $\mathbf{A}_{PCP}$ formed by the meta-path $PCP$ is $\beta_2$. Then, the second-order product of the meta-paths $PAP$ and $PCP$ will contain the four meta-paths $PAPAP$, $PAPCP$, $PCPAP$, and $PCPCP$: 


\begin{equation}
\begin{aligned}
\label{eq:more_path}
\mathbb{A}^2=(\beta_1 \mathbf{A}_{PAP}+\beta_2\mathbf{A}_{PCP})^2&=\beta_1\beta_1\mathbf{A}_{PAPAP}+\beta_1\beta_2\mathbf{A}_{PAPCP}\\
&+\beta_2\beta_1\mathbf{A}_{PCPAP}+\beta_2\beta_2\mathbf{A}_{PCPCP},
\end{aligned}
\end{equation}
where the adjacency matrix of a meta-path can be written as the product of the adjacency matrices of multiple meta-paths, for example,  $\mathbf{A}_{PAPCP} =\mathbf{A}_{PAP} \cdot \mathbf{A}_{PCP} $.
\subsection{Training Objective}

After obtaining the local independent filtering representation $\mathbf{Z}_{l}$ and the global hybrid filtering representation $\mathbf{Z}_{g}$, we add them together to get the final representation $\mathbf{Z}$:

\begin{equation}
\mathbf{Z} = \mathbf{Z}_{l} + \mathbf{Z}_{g}.
\end{equation}



We adopt a multi-layer perceptron (MLP) with parameter $\theta$ to predict the label distribution:
\begin{equation}
\hat{\mathbf{y}}=\operatorname{MLP}\left(\mathbf{Z} ; \theta\right),
\end{equation}
where $\hat{\mathbf{y}}$ is the predicted label distribution. Then, we optimize
the cross-entropy loss of the node $j$:
\begin{equation}
\mathcal{L}=\sum_{j \in \mathcal{V}_{\text {train }}} \operatorname{CrossEntropy}\left(\hat{\mathbf{y}}^j, \mathbf{y}^j\right),
\end{equation}
where $\mathcal{V}_{\text {train }}$ is the training node set, and $\mathbf{y}^j$
is the ground-truth one-hot label vector of node $j$.



\subsection{Model Analysis}
\begin{table}[t]

\centering
\caption{Comparison of the number of parameters and terms in the filter part between H$^2$SGNN and PSHGCN, where $K$ is the order and $R$ is the number of relationship matrices.}
\resizebox{0.48\textwidth}{!}{
\begin{tabular}{lccc}
\toprule
Method &Number of parameters & Number of items \\ \midrule
 
PSHGCN \cite{PSHGCN} & $\frac{R^{K+1}-1}{R-1}$ & $\frac{R^{K+1}-1}{R-1}$\\
H$^2$SGNN-$local$ &   $R(K+1)$ & $R(K+1)$\\
H$^2$SGNN-$global$ &   $R+K+1$ & $K+1$\\
H$^2$SGNN &   $(R+1)(K+1)+R$ & $(R+1)(K+1)$\\
\bottomrule
\end{tabular}
}
\label{tab:item_para}
\end{table}

\label{sec:analysis}

PSHGCN \cite{PSHGCN} has utilized multivariate non-commutative polynomials for convolution on heterogeneous graphs, which assert their capability to fit arbitrary filter functions. For instance, a second-order polynomial \( h \) with three variables can be expressed as:

 


\begin{equation}
\begin{aligned}
h(\mathbf{A}_1, \mathbf{A}_2, \mathbf{A}_3)
&= w_0 \mathbf{I} + w_1 \mathbf{A}_1 + w_2 \mathbf{A}_2 + w_3 \mathbf{A}_3\\
&+ w_{1,1} \mathbf{A}_1 \mathbf{A}_1 + w_{1,2} \mathbf{A}_1 \mathbf{A}_2 + w_{1,3} \mathbf{A}_1 \mathbf{A}_3 \\
&+ w_{2,1} \mathbf{A}_2 \mathbf{A}_1 + w_{2,2} \mathbf{A}_2 \mathbf{A}_2 + w_{2,3} \mathbf{A}_2 \mathbf{A}_3 \\
&+ w_{3,1} \mathbf{A}_3 \mathbf{A}_1 + w_{3,2} \mathbf{A}_3 \mathbf{A}_2 + w_{3,3} \mathbf{A}_3 \mathbf{A}_3 ,
\end{aligned}
\label{eq:multivariate}
\end{equation}
where $w$ are learnable parameters, and $\mathbf{A}_1$, $\mathbf{A}_2$ and $\mathbf{A}_3$ are different relationship matrices. Despite their high expressive power, these models incur a significant increase in memory and parameter count, since the number of terms grows exponentially with the polynomial order \( K \), i.e., $\frac{R^{K+1}-1}{R-1}$. Conversely, as shown in Table \ref{tab:item_para}, our proposed model achieves linear item and parameter growth, while approaching the expressive power of PSHGCN \cite{PSHGCN}. The following proposition offers a theoretical foundation:

\begin{Proposition}
\textbf{The $n$-order terms in the global hybrid filter correspond to terms in the multivariate non-commutative polynomial with $n$ matrix products.}
\end{Proposition}


We provide the proof of this proposition in Appendix A. Let's take an intuitive example. When the order $K=2$ and the number of relationship matrices $R=3$, The second order of the global adjacency matrix $\mathbb{A}$ in Eq. (\ref{eq:beta}) can be written as follows:

\begin{equation}
\begin{aligned}
\mathbb{A}^2 &= (\beta_{1}\mathbf{A}_{1}+\beta_{2}\mathbf{A}_{2}+\beta_{3}\mathbf{A}_{3})^2 \\
&=\beta_{1}\beta_{1}\mathbf{A}_{1}\mathbf{A}_{1}+\beta_{1}\beta_{2}\mathbf{A}_{1}\mathbf{A}_{2} +\beta_{1}\beta_{3}\mathbf{A}_{1}\mathbf{A}_{3}\\
&+\beta_{2}\beta_{1}\mathbf{A}_{2}\mathbf{A}_{1}+\beta_{2}\beta_{2}\mathbf{A}_{2}\mathbf{A}_{2} +\beta_{2}\beta_{3}\mathbf{A}_{2}\mathbf{A}_{3} \\
&+\beta_{3}\beta_{1}\mathbf{A}_{3}\mathbf{A}_{1}+\beta_{3}\beta_{2}\mathbf{A}_{3}\mathbf{A}_{2}+\beta_{3}\beta_{3}\mathbf{A}_{3}\mathbf{A}_{3},
\end{aligned}
\end{equation}
where $\beta_{1}$, $\beta_{2}$ and $\beta_{3}$ are learnable parameters. It can be seen that it corresponds exactly to the last nine terms in the multivariate non-commutative polynomial $h\left(\mathbf{A}_1, \mathbf{A}_2, \mathbf{A}_3\right)$ in Eq. (\ref{eq:multivariate}). Therefore, a multivariate non-commutative polynomial can actually be composed of polynomials with multivariate sums as independent variables, which can reduce parameters and memory consumption. Hence, the proposed H$^2$SGNN model can improve the expressiveness by increasing the order $K$ due to lower resource consumption. In addition, compared to PSHGCN, we add local individual filters to filter each meta-path separately, which explicitly filters each meta-path and enhances interpretability.


\section{Experiment}
In this section, to fully evaluate the performance of the proposed H$^2$SGNN model, we present a series of comprehensive experiments to answer the following research questions (\textbf{RQ}s): 
\begin{itemize}[leftmargin=0.5cm, itemindent=0cm]
\item \textbf{RQ1}: How does H$^2$SGNN perform compared to existing state-of-the-art methods? 
\item \textbf{RQ2}: Do both local independent filters and global hybrid filters help improve performance? 
\item \textbf{RQ3}: Are higher-order neighbors beneficial to model performance?
\item \textbf{RQ4}: As the order $K$ grows, is the proposed H$^2$SGNN efficient in terms of memory and parameters?
\item \textbf{RQ5}: Can the learned filters enhance the interpretability of spectral GNNs on heterogeneous graphs?

\end{itemize}
More experimental results, such as sensitivity analysis on different 
polynomial basis and importance analysis of each meta-path can be found in Appendix E.
\begin{table}[t]
\centering
    \caption{Dataset Statistics.}
    
    \begin{tabular}{lccccccc}
    \toprule
      Dataset & Nodes & \makecell[c]{Node \\ Types} & Edges  & Target & Categories  \\
    \midrule
       DBLP & 26,128 & 4 & 239,566  & author & 4 \\
       ACM & 10,942 & 4 & 547,872  & paper & 3 \\
       IMDB & 24,420 & 4 & 86,642  & movie & 5  \\
       AMiner & 55,783 & 3 & 153,676  & paper & 4  \\
    \bottomrule
    \end{tabular}
    \label{tb:statistics}
\end{table}

\subsection{Experimental Setup}
\textbf{Datasets.}
We evaluate the proposed H$^2$SGNN model on semi-supervised node classification tasks on four widely used heterogeneous graph datasets, including three academic citation heterogeneous graphs DBLP~\cite{hgb}, ACM~\cite{hgb} and AMiner~\cite{HeCo}, a movie rating graph IMDB~\cite{hgb}. The statistical information of these datasets, such as the number of nodes and edges, the target node type, and the number of categories, is shown in Table \ref{tb:statistics}. For detailed hyperparameters and meta-path settings, please see Appendix D.1 and D.2.

\begin{table*}[t]
\centering
\small
\caption{Node classification performance (Mean F1 scores $\pm$ standard errors) comparison on four datasets, where ”-” indicates results not available in the original paper.}
\resizebox{\textwidth}{!}{
\begin{tabular}{llcccccccccc}
\toprule
        \multirow{3}{*}{\thead{Method\\
        Group}} & \multirow{3}{*}{Method} & \multicolumn{2}{c}{DBLP} & \multicolumn{2}{c}{ACM} & \multicolumn{2}{c}{IMDB} & \multicolumn{2}{c}{AMiner} 
       \\ 
       \cmidrule{3-10}
      &  &Macro-F1 &Micro-F1 &Macro-F1 &Micro-F1 &Macro-F1 &Micro-F1 &Macro-F1 &Micro-F1 
       \\ \midrule
  \multirow{2}{*}{{\thead{Homogeneous \\ Homophilic}}} &  
GCN \cite{GCN}   &90.84$_{\pm\text{0.32}}$ &91.47$_{\pm\text{0.34}}$ & 92.17$_{\pm\text{0.24}}$ & 92.12$_{\pm\text{0.23}}$ & 62.37$_{\pm\text{1.35}}$ & 68.13$_{\pm\text{0.83}}$  &75.63$_{\pm\text{1.08}}$ &85.77$_{\pm\text{0.43}}$
\\
& GAT \cite{GAT}   &93.83$_{\pm\text{0.27}}$ &93.39$_{\pm\text{0.30}}$ & 92.26$_{\pm\text{0.94}}$ & 92.19$_{\pm\text{0.93}}$ & 62.45$_{\pm\text{1.36}}$ & 68.08$_{\pm\text{0.49}}$  &75.23$_{\pm\text{0.60}}$ &85.56$_{\pm\text{0.65}}$
\\
\midrule
\multirow{2}{*}{{\thead{Homogeneous \\ Heterophilic}}} & 
GPRGNN \cite{GPR-GNN}   &91.66$_{\pm\text{1.01}}$ &92.45$_{\pm\text{0.76}}$ & 92.36$_{\pm\text{0.28}}$ & 92.28$_{\pm\text{0.27}}$ & 63.02$_{\pm\text{1.48}}$ & 68.83$_{\pm\text{0.95}}$ &75.32$_{\pm\text{0.67}}$ &86.13$_{\pm\text{0.58}}$
\\
& ChebNetII \cite{ChebNetII} &92.05$_{\pm\text{0.53}}$ &92.97$_{\pm\text{0.48}}$ & 92.45$_{\pm\text{0.37}}$ & 92.33$_{\pm\text{0.38}}$ & 62.54$_{\pm\text{1.29}}$ & 68.33$_{\pm\text{0.92}}$ &75.59$_{\pm\text{0.73}}$ &85.82$_{\pm\text{0.52}}$
\\ \midrule
\multirow{8}{*}{{\thead{Heterogeneous \\
Homophilic}}} & 
RGCN \cite{RGCN}   &91.52$_{\pm\text{0.50}}$  &92.07$_{\pm\text{0.50}}$ & 91.55$_{\pm\text{0.74}}$ & 91.41$_{\pm\text{0.75}}$ & 63.24$_{\pm\text{0.57}}$ & 66.51$_{\pm\text{0.28}}$  &63.03$_{\pm\text{2.27}}$ &82.79$_{\pm\text{1.12}}$\\

 &HAN \cite{HAN}   &91.67$_{\pm\text{0.49}}$ &92.05$_{\pm\text{0.62}}$ & 90.89$_{\pm\text{0.43}}$ & 90.79$_{\pm\text{0.43}}$ & 62.05$_{\pm\text{0.93}}$ & 67.69$_{\pm\text{0.64}}$ &63.86$_{\pm\text{2.15}}$ &82.95$_{\pm\text{1.33}}$ \\

 &GTN \cite{gtn}   &93.52$_{\pm\text{0.55}}$ &93.97$_{\pm\text{0.54}}$ & 91.31$_{\pm\text{0.70}}$ & 91.20$_{\pm\text{0.71}}$ & 64.59$_{\pm\text{1.03}}$ & 68.27$_{\pm\text{0.65}}$ &72.39$_{\pm\text{1.79}}$ &84.74$_{\pm\text{1.24}}$ 
\\
 &MAGNN \cite{magnn} &93.28$_{\pm\text{0.51}}$ &93.76$_{\pm\text{0.45}}$ & 90.88$_{\pm\text{0.64}}$ & 90.77$_{\pm\text{0.65}}$ & 61.36$_{\pm\text{2.85}}$ & 67.82$_{\pm\text{1.54}}$ &71.56$_{\pm\text{1.63}}$ &83.48$_{\pm\text{1.37}}$ \\

 &EMRGNN \cite{emrgnn} &92.19$_{\pm\text{0.38}}$ &92.57$_{\pm\text{0.37}}$ & 92.93$_{\pm\text{0.34}}$ & 93.85$_{\pm\text{0.33}}$ & 65.63$_{\pm\text{1.97}}$ & 68.76$_{\pm\text{0.78}}$ &73.74$_{\pm\text{1.25}}$ &85.46$_{\pm\text{0.74}}$\\

 &MHGCN \cite{mhgcn} &93.56$_{\pm\text{0.41}}$ &94.03$_{\pm\text{0.43}}$ & 92.12$_{\pm\text{0.66}}$ & 91.97$_{\pm\text{0.68}}$ & 67.59$_{\pm\text{1.25}}$ & 70.28$_{\pm\text{0.71}}$ &73.56$_{\pm\text{1.75}}$ &85.18$_{\pm\text{1.28}}$\\

 &SimpleHGN  \cite{hgb} &94.01$_{\pm\text{0.24}}$ &94.46$_{\pm\text{0.22}}$ & 93.42$_{\pm\text{0.44}}$ & 93.35$_{\pm\text{0.45}}$ & 68.72$_{\pm\text{1.54}}$ & 70.83$_{\pm\text{1.07}}$ &75.43$_{\pm\text{0.88}}$ &86.52$_{\pm\text{0.73}}$ 
\\

 &HALO \cite{halo}  &92.37$_{\pm\text{0.32}}$ &92.84$_{\pm\text{0.34}}$ & 93.05$_{\pm\text{0.31}}$ & 92.96$_{\pm\text{0.33}}$ & 71.63$_{\pm\text{0.77}}$ &73.81$_{\pm\text{0.72}}$ &74.91$_{\pm\text{1.23}}$
&87.25$_{\pm\text{0.89}}$ 
\\
 &SeHGNN~\cite{sehgnn} &95.06$_{\pm\text{0.17}}$ &95.42$_{\pm\text{0.17}}$ & 94.05$_{\pm\text{0.35}}$ & 93.98$_{\pm\text{0.36}}$ & 
71.71$_{\pm\text{0.62}}$ & 73.42$_{\pm\text{0.47}}$
&76.83$_{\pm\text{0.57}}$
&86.96$_{\pm\text{0.64}}$
\\

\midrule
 \multirow{4}{*}{{\thead{Heterogeneous \\
Heterophilic}}} &  
HDHGR \cite{HDHGR} & \text{94.43}$_{\pm\text{0.20}}$ &\text{94.73}$_{\pm\text{0.16}}$ & \text{93.88}$_{\pm\text{0.20}}$ & \text{93.80}$_{\pm\text{0.20}}$ & \text{58.97}$_{\pm\text{0.58}}$ & \text{59.32}$_{\pm\text{0.53}}$ &--
&-- \\

& Hetero2Net \cite{Hetero2Net}& \text{94.03}$_{\pm\text{0.35}}$ &\text{94.46}$_{\pm\text{0.37}}$ & \text{94.01}$_{\pm\text{0.54}}$ & \text{93.91}$_{\pm\text{0.61}}$ & \text{65.37}$_{\pm\text{0.48}}$ & \text{69.61}$_{\pm\text{0.72}}$ & --
&-- \\
& PSHGCN \cite{PSHGCN}&\textbf{95.27}$_{\pm\textbf{0.13}}$ &\textbf{95.61}$_{\pm\textbf{0.12}}$ & \underline{\text{94.35}$_{\pm\text{0.23}}$} & 
\underline{\text{94.27}$_{\pm\text{0.23}}$} & \underline{\text{72.33}$_{\pm\text{0.57}}$}& 
\underline{\text{74.46}$_{\pm\text{0.32}}$} & 
\underline{\text{77.26}$_{\pm\text{0.75}}$} 
& \underline{\text{88.21}$_{\pm\text{0.31}}$} 
\\

& \textbf{H$^2$SGNN (ours)}& \underline{\text{95.19}$_{\pm\text{0.11}}$} &\underline{\text{95.56}$_{\pm\text{0.11}}$} & \textbf{94.47}$_{\pm\textbf{0.25}}$ & \textbf{94.38}$_{\pm\textbf{0.26}}$ & \textbf{73.04}$_{\pm\textbf{0.65}}$  & \textbf{75.46}$_{\pm\textbf{0.43}}$  &\textbf{78.44}$_{\pm\textbf{1.10}}$ & \textbf{88.53}$_{\pm\textbf{0.65}}$  \\

\bottomrule
\end{tabular}}
\label{tb:main_result}
\end{table*}

\textbf{Settings.} 
For a fair comparison, we adopt the experimental settings used in the Heterogeneous Graph Benchmark (HGB)~\cite{hgb}, and follow its standard split with training/validation/test sets accounting for 24\%/6\%/70\% respectively.  We use Micro-F1 and Macro-F1 metrics as evaluation indicators. All experiments are performed five times, and we report the average results and their corresponding standard errors. All experiments are conducted on a machine with 3 NVIDIA A5000 24GB GPUs and Intel(R) Xeon(R) Silver 4310 2.10GHz CPU. 


\textbf{Baselines. }
In order to fully verify the performance of the proposed H$^2$SGNN, we select four types of baseline methods, namely homogeneous homophilic GNN \cite{GCN,GAT}, homogeneous heterophilic GNN \cite{GPR-GNN,ChebNetII}, heterogeneous homophilic GNN \cite{RGCN,HAN,gtn,magnn,emrgnn,mhgcn,hgb,halo,sehgnn}, and heterogeneous heterophilic GNN \cite{HDHGR,Hetero2Net,PSHGCN}. Appendix D.3 shows the details of these baseline methods.

\subsection{Main Results (\textbf{RQ1})}

Table \ref{tb:main_result} presents the experimental results on four datasets, with the first two results highlighted in bold and underlined, respectively. For HDHGR and Hetero2Net, we use the results from their original papers. For other baselines, we directly quote the reported results from PSHGCN \cite{PSHGCN}. From Table \ref{tb:main_result}, we have the following observations: 1) Heterogeneous heterophilic GNNs, such as HDHGR \cite{HDHGR}, Hetero2Net \cite{Hetero2Net}, and PSHGCN \cite{PSHGCN} \footnote{Although PSHGCN does not explicitly address heterophily, we still classify it as a Heterogeneous Heterophilic GNN because its designed filters can fit arbitrary functions.}, achieve competitive performance. This indicates that addressing heterophily can improve performance. 2) The proposed H$^2$SGNN model outperforms all baselines on three out of four datasets, except for DBLP. This highlights the effectiveness of H$^2$SGNN in handling heterogeneous heterophilic graphs. 3) Although PSHGCN claims to have stronger expressive power, compared with PSHGCN, the proposed H$^2$SGNN achieves an average improvement of 1.2\% and 0.9\% on the IMDB and AMiner datasets, respectively, and also achieves competitive performance on the ACM and DBLP datasets. This shows that the proposed H$^2$SGNN can exploit high-order neighbors to learn more meta-paths, thereby improving the performance.

\begin{figure}[t]   
	\centering 
	\includegraphics[width=\linewidth]{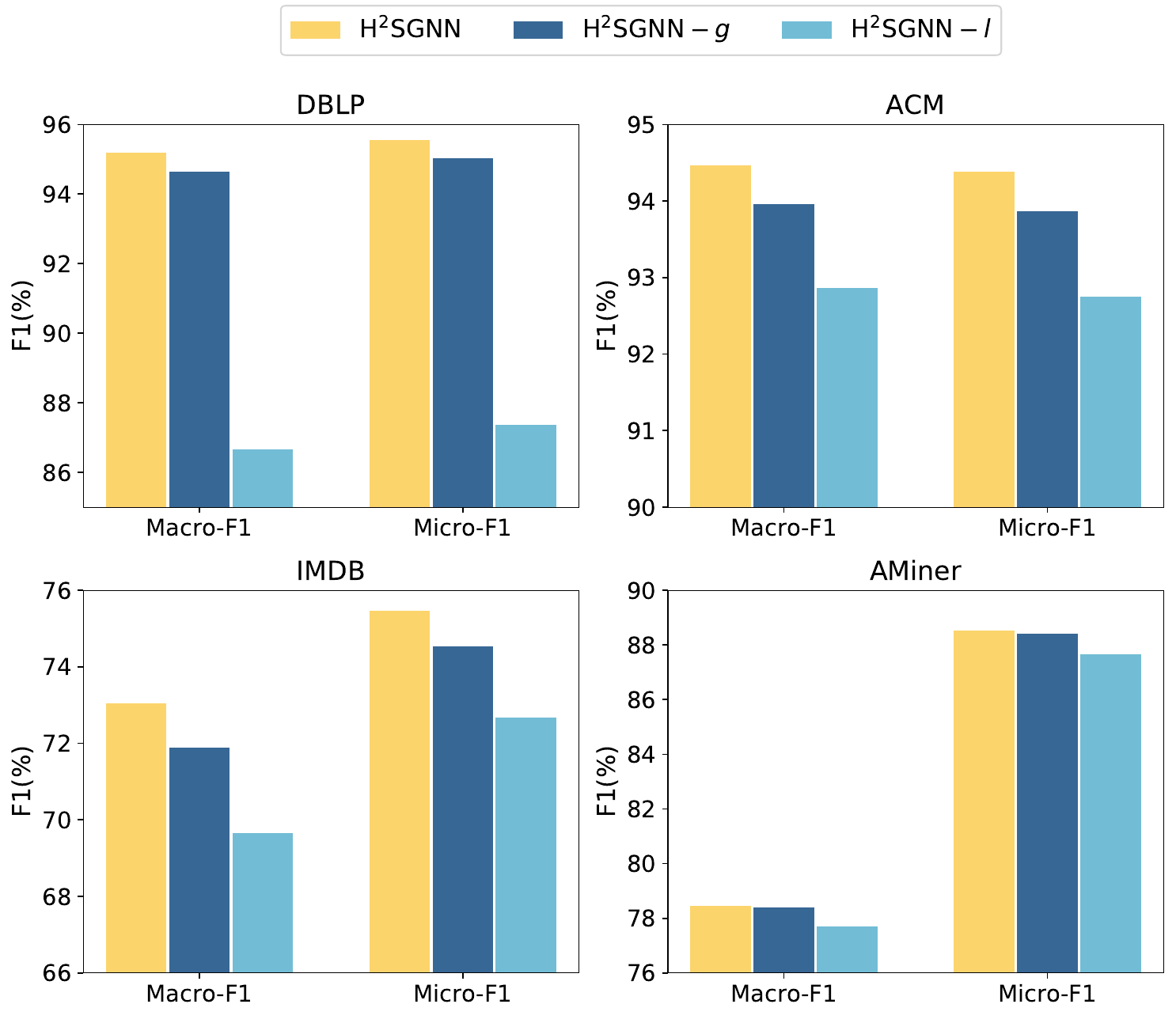}
	\caption{Ablation study of proposed H$^2$SGNN on four datasets with two variants H$^2$SGNN-$g$ and H$^2$SGNN-$l$.}
 \label{fig:ablation}
\end{figure}

\subsection{Ablation Analysis (\textbf{RQ2})}
\label{sec:abl}
This subsection aims to verify the usefulness of local independent filtering and global hybrid filtering through ablation analysis. We believe that neither the local independent filtering nor the global hybrid filtering is optimal, and only by combining them can the optimal performance be achieved. Therefore, we design two variants H$^2$SGNN-$g$ and H$^2$SGNN-$l$ to verify our conjecture. 
\begin{itemize}[leftmargin=0.5cm, itemindent=0cm]
\item  The variant H$^2$SGNN-$g$ uses only the global hybrid filtering, removing the local independent filtering.
\item The variant H$^2$SGNN-$l$ uses only the local independent filtering, removing the global hybrid filtering. 
\end{itemize}

Figure \ref{fig:ablation} presents the results of ablation experiments on four datasets. Based on these results, we have the following observations: 1) Utilizing solely the global hybrid filtering H$^2$SGNN-$g$ results in diminished performance. This decline occurs because the absence of local independent filtering impedes the model's ability to learn node representations independently under diverse homophily of different meta-paths. 2) Relying exclusively on the local independent filtering H$^2$SGNN-$l$ also leads to suboptimal performance. In this case, the H$^2$SGNN-$l$ model cannot learn the interactions between different meta-paths and more possible meta-paths. 3) 
The performance of H$^2$SGNN-$g$ is second only to H$^2$SGNN, which shows the strong competitiveness of global hybrid filtering, thanks to extensive meta-path exploration. Combined with local independent filtering that considers diverse homophily of different meta-paths, the model performance is further improved.
\begin{figure}[t]   
	\centering 
	\includegraphics[width=\linewidth]{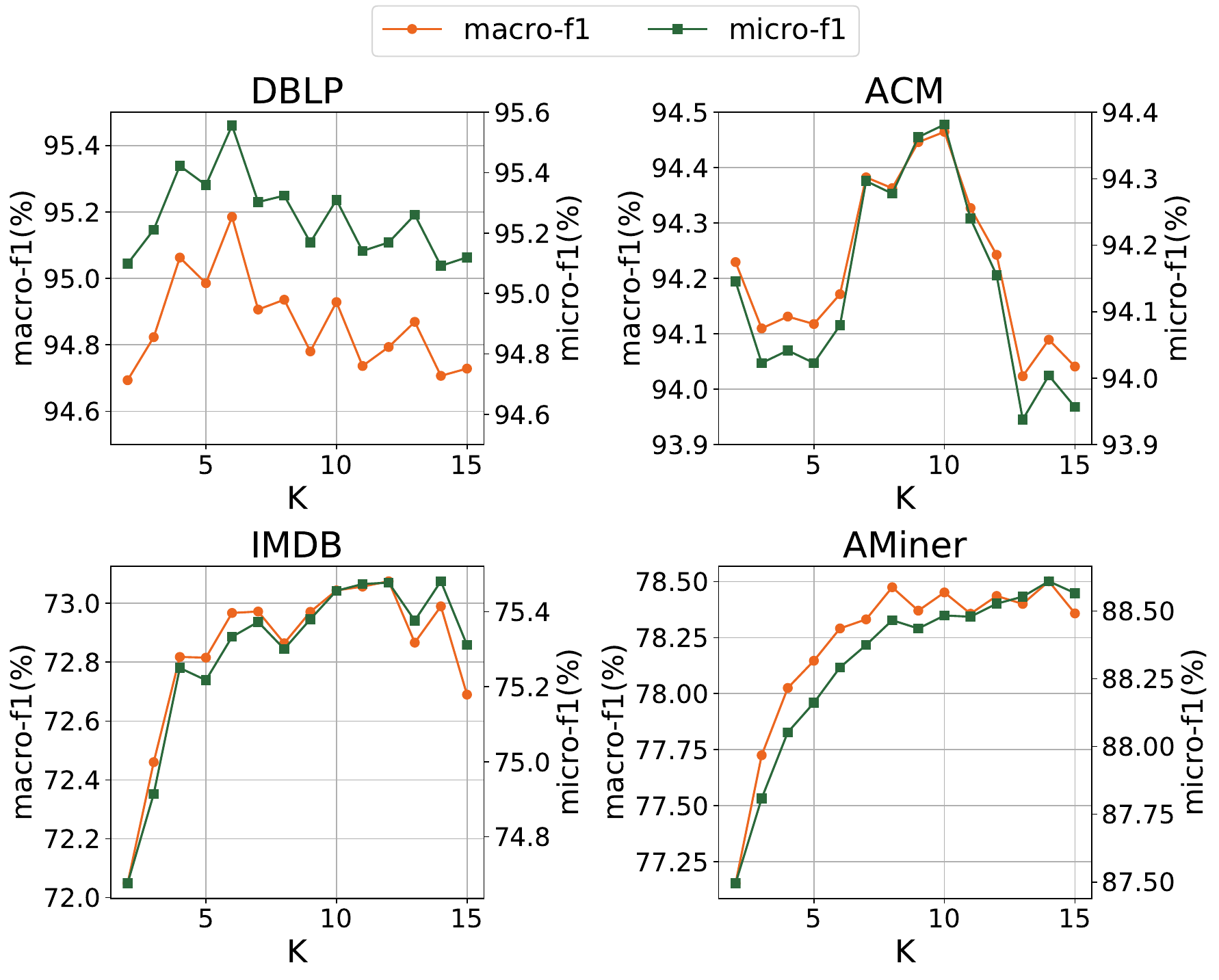}
	\caption{Effect of order $K$ on model performance.}
  \label{fig:hyperparameter}
\end{figure}
\subsection{Parameter Sensitivity Analysis (\textbf{RQ3})}



It is well understood that the order of the polynomial filter significantly impacts model performance. Therefore, we analyze the effect of the order $K$ of the polynomial on model performance for four datasets. As illustrated in Figure \ref{fig:hyperparameter}, for the DBLP and ACM datasets, increasing the order $K$ initially enhances model performance by expanding the neighborhood range. However, beyond a certain threshold, further increasing the order $K$ leads to a downward trend in model performance, which may be due to the introduction of irrelevant noise. On the contrary, for the IMDB and AMiner datasets, model performance generally improves with higher values of $K$, indicating that incorporating more neighbor information is beneficial for DBLP and ACM datasets. Hence, compared to PSHGCN \cite{PSHGCN}, proposed H$^2$SGNN can achieve better performance with a higher order $K$. This is because we can increase the order $K$ to achieve better performance due to the linear parameters and memory consumption of the proposed H$^2$SGNN. In contrast, PSHGCN is difficult to scale to higher order $K$ due to its exponentially increasing linear parameters and memory consumption. As a consequence, by utilizing more high-order neighbors to learn more meta-paths, the proposed H$^2$SGNN achieves more significant improvements compared to PSHGCN on IMDB and AMiner datasets.

%




\begin{figure}[t]   
	\centering 
	\includegraphics[width=\linewidth]{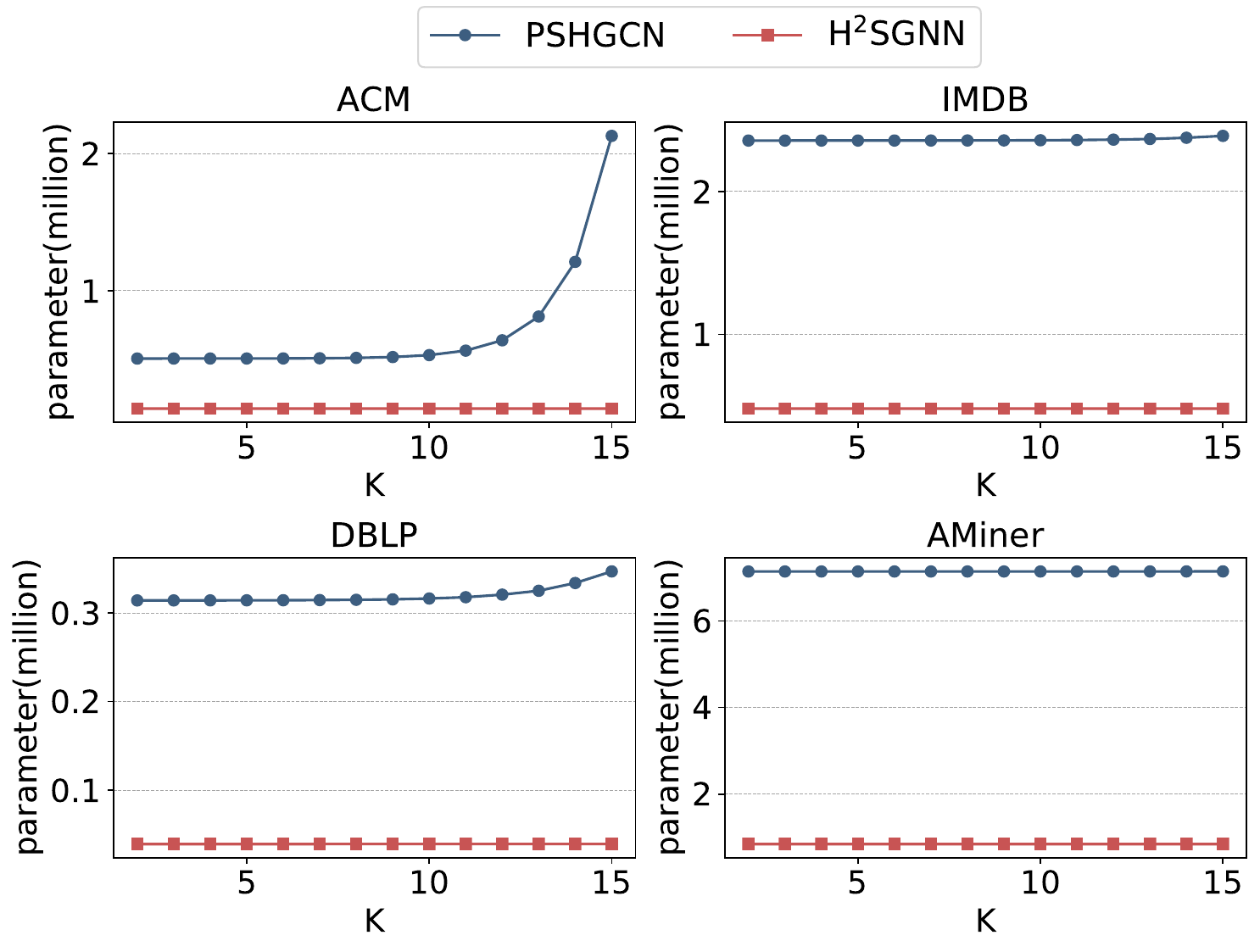}
	\caption{Parameter usage comparison of H$^2$SGNN and PSHGCN under different order $K$.}
  \label{fig:parameter}
\end{figure}

\begin{figure}[t]   
	\centering 
	\includegraphics[width=\linewidth]{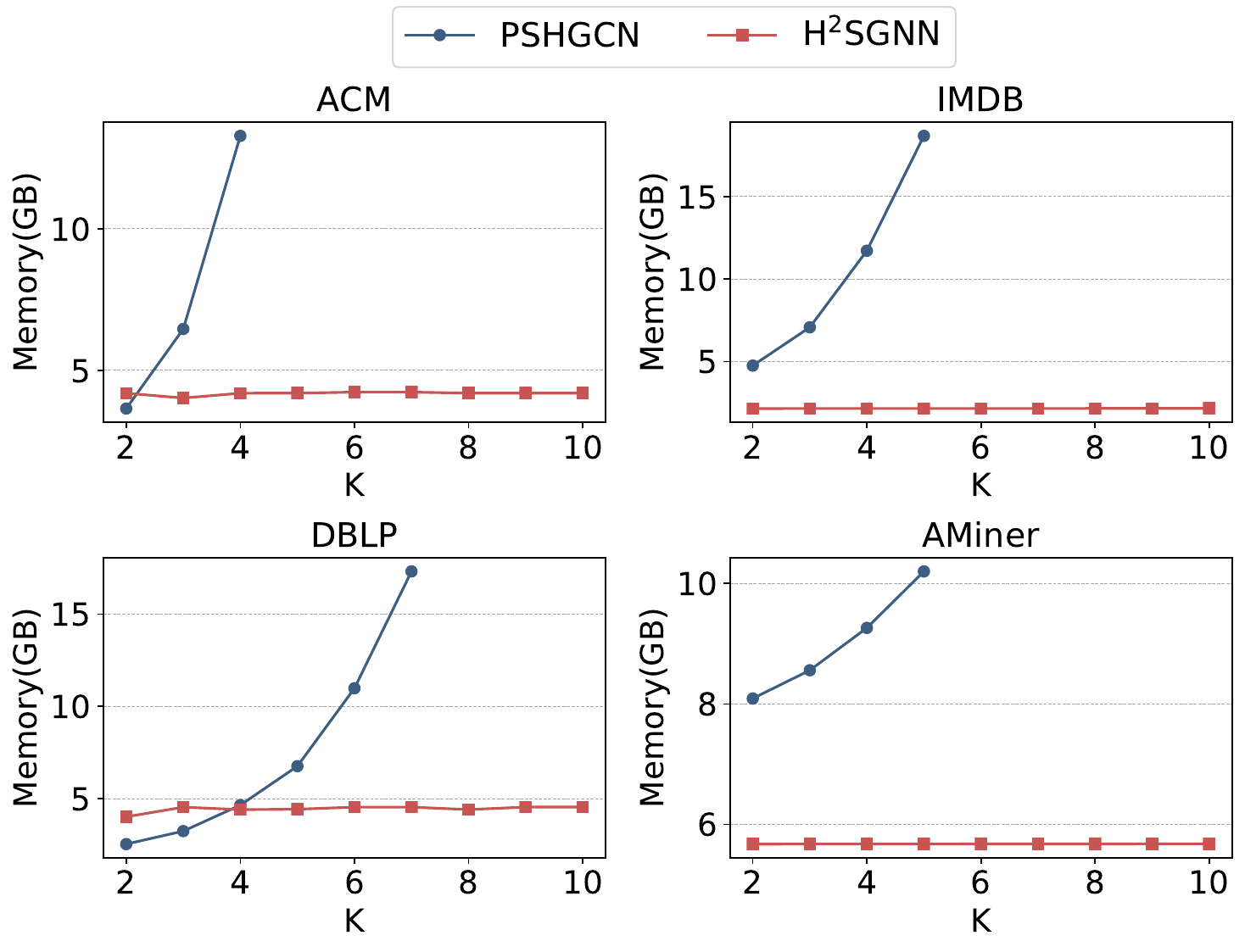}
	\caption{Memory consumption comparison under different order $K$, where missing curves represent instances where memory exceeded the 24GB limit of the GPU, resulting in out-of-memory (OOM) errors.}
  \label{fig:memory}
\end{figure}

\subsection{Efficiency Studies (\textbf{RQ4})}
\label{efficiency}

In this subsection, we evaluate the advantages of the proposed H$^2$SGNN in terms of memory and parameter usage\footnote{For a fair comparison, PSHGCN and the proposed H$^2$SGNN adopt the same multi-layer perceptron (MLP) for efficiency analysis.}. Figure \ref{fig:parameter} compares the parameter usage of H$^2$SGNN and the most competitive model PSHGCN \cite{PSHGCN}, under different orders $K$. It reveals that PSHGCN has several times more parameters than the proposed H$^2$SGNN on each dataset, which is mainly because H$^2$SGNN only focuses on learning the representation of target nodes, while PSHGCN learns the representation of all types of nodes. When the number of other types of nodes is much larger than the number of target node types, the advantage of H$^2$SGNN will be more obvious. Moreover, as analyzed in subsection \ref{sec:analysis}, due to the exponential growth of the number of items: $\frac{R^{K+1}-1}{R-1}$, the number of parameters of PSHGCN on the ACM and DBLP dataset also increases exponentially, while the number of parameters of H$^2$SGNN always remains low, due to the linear parameters $(R+1)(K+1)+R$. 


In addition, to show the computational cost requirements of PSHGCN and the proposed H$^2$SGNN model, we also conduct memory consumption experiments. As illustrated in Figure 7, the memory consumption of PSHGCN \cite{PSHGCN} grows exponentially with the order $K$. This is due to the exponential increase in the number of expansion terms in the multivariate polynomial as $K$ rises: $\frac{R^{K+1}-1}{R-1}$. As the number of terms increases, the model requires more memory to store the additional items. This result is consistent with our complexity analysis in subsection 3.4. Therefore, when $K$ exceeds 7, PSHGCN \cite{PSHGCN} encounters an out-of-memory (OOM) error in 24GB GPU. In contrast, as shown in Table \ref{tab:item_para}, thanks to the linear items $(R+1)(K+1)$, the memory consumption of H$^2$SGNN is always low. Therefore, the proposed H$^2$SGNN model can utilize higher-order neighbors to improve performance without large memory consumption.

\begin{figure}[t]   
	\centering 
	\includegraphics[width=\linewidth]{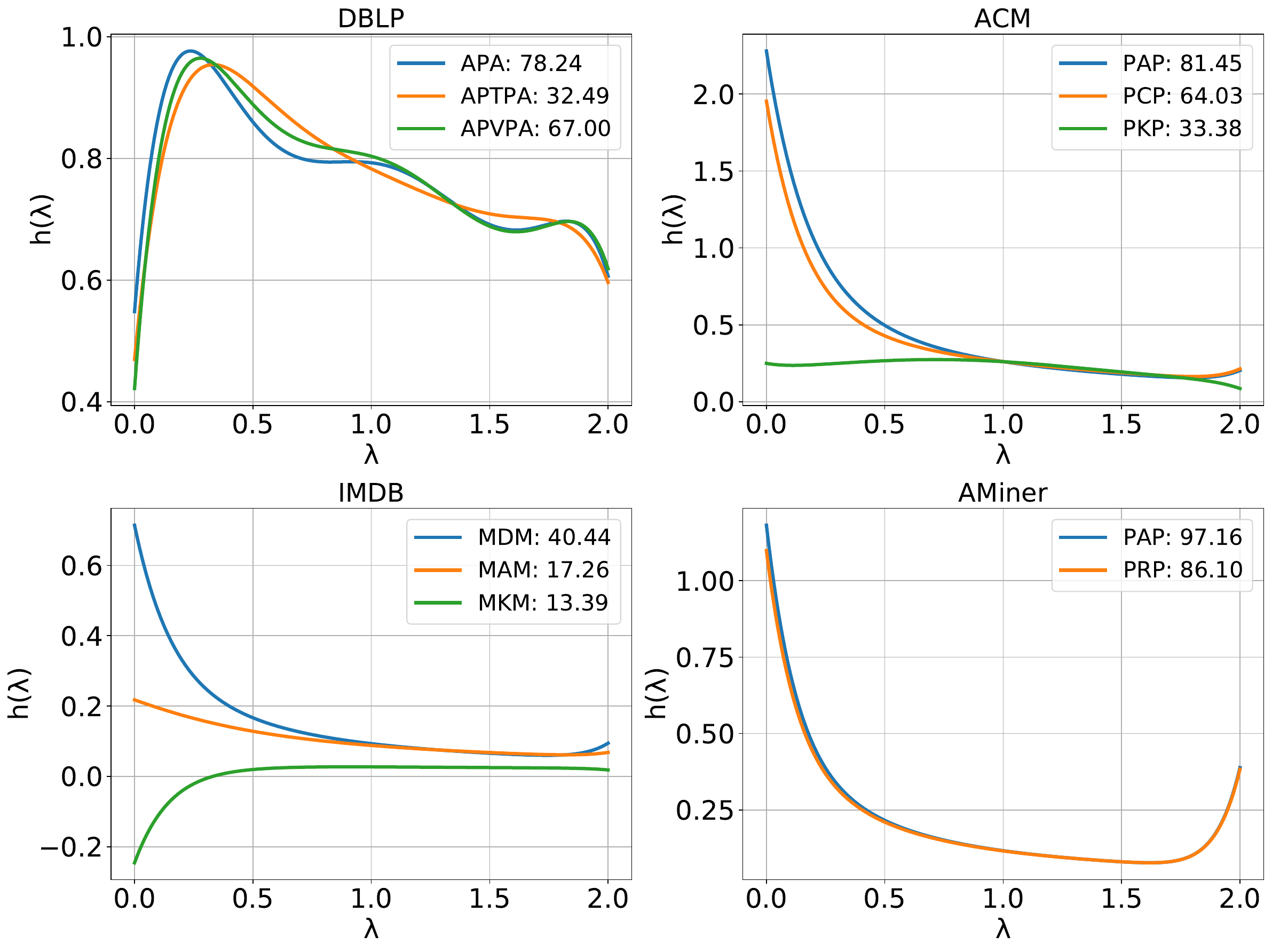}
	\caption{Filter visualization. The horizontal axis is the eigenvalue of normalized Laplacian matrix, and the vertical axis is the coefficient of the corresponding eigenvalue learned by the model. The legend shows different meta-paths and their corresponding homophily (\%).}
   \label{fig:filter}
\end{figure}

\subsection{Filter Visualization (\textbf{RQ5})}

In subsection \ref{local_individual_filtering}, we add local independent filtering to apply different filters to meta-paths of different homophily. To verify the filtering effect of local independent filtering on meta-paths of different homophily, we visualize the learned filters by different meta-paths on four datasets. From Figure \ref{fig:filter}, we can see that the filters learned by different meta-paths under the same dataset may be similar, such as the DBLP and AMiner datasets, which may be due to their similar homophily. However, the filters learned by different meta-paths under the same dataset may also be different. For example, both the $PAP$ and $PCP$ meta-paths of the ACM dataset learn low-pass filters due to their high homophily, while $PKP$ learns a flatter filter due to its lower homophily. In addition, in the IMDB dataset, unlike the $MDM$ and $MKM$ meta-paths, the $MAM$ meta-path also learns a flatter filter. In summary, filter visualization for the four datasets indicates that it is necessary to apply different filters to different meta-paths, which corresponds to our proposed local independent filtering module.

Moreover, the filter visualization in Figure \ref{fig:filter} enhances the interpretability of spectral GNN on heterogeneous graphs, because the learned filters can implicitly reflect the homophily of the graph, for example, the learned low-pass filter represents a homophilic graph. 
At the same time, if the homophily of a meta-path subgraph can be pre-calculated or estimated, a filter for this meta-path can be designed in advance to perform personalized filtering \cite{NewtonNet}.


\section{Conclusion and Future Work}
In this paper, to solve the two challenges of heterogeneity and heterophily, we proposed a heterogeneous heterophilic spectral graph neural network (H$^2$SGNN), which consists of two modules: local independent filtering and global hybrid filtering. Local independent filtering aims to learn node representations of meta-paths under different homophily. Global hybrid filtering exploits high-order neighbors to learn more possible meta-paths. Moreover, The proposed H$^2$SGNN enhances the interpretability of spectral GNN on heterogeneous graphs, because the learned filters can reflect the homophily. Extensive experiments demonstrate the superior performance of H$^2$SGNN due to its less memory and parameter consumption.


Even though the proposed H$^2$SGNN can automatically learn more meta-paths, a few meta-paths still need to be specified in advance. Therefore, how to apply spectral GNNs in heterogeneous graphs without defining any meta-paths is a problem worth studying. In addition, since our model 
exhibits significant parameter and memory advantages especially when the order of the graph grows, it is promising to extend our method to more scenarios with higher-order graphs.

\bibliographystyle{ACM-Reference-Format}
\bibliography{sample-base}
\newpage

\appendix

\newpage
\section{Proof of Proposition 1}
For convenience, we omit the coefficient $\beta_i$ of each meta-path and assume global adjacency $\mathbb{A}=\sum_{i=1}^{R}\mathbf{A}_i$. Then 
the $k$-th order global hybrid filter with $R$ meta-paths can be written as:

\begin{equation}
\begin{aligned}
\mathbb{A}^k &= (\mathbf{A}_{1}+\mathbf{A}_{2}+...+\mathbf{A}_{R})^k \\
&=\sum_{i_1, i_2, \ldots, i_k=1}^R A_{i_1} A_{i_2} \ldots A_{i_k}.
\end{aligned}
\label{eq:Multivariate_Functions_sum}
\end{equation}
In MGNN \cite{MGNN} and PSHGCN \cite{PSHGCN}, we define all terms of $k$-matrix products as $h(\mathbf{A}_{1},\mathbf{A}_{2},...,\mathbf{A}_{R})_k$ , which are all products of $k$ matrices selected from matrices $\mathbf{A}_{1},\mathbf{A}_{2},...,\mathbf{A}_{R}$ in order. Then, we have:
\begin{equation}
\begin{aligned}
h(\mathbf{A}_{1},\mathbf{A}_{2},...,\mathbf{A}_{R})_k = \sum_{i_1, i_2, \ldots, i_k=1}^R A_{i_1} A_{i_2} \ldots A_{i_k}.
\end{aligned}
\label{eq:Multivariate_Functions}
\end{equation}
Eq. (\ref{eq:Multivariate_Functions_sum}) and Eq. (\ref{eq:Multivariate_Functions}) are exactly the same, because their sum contains the expanded terms for all possible matrix product orders. This completes the proof of Proposition 1 in main paper.

\section{The Overall Training Process}
\begin{algorithm}[t] 
		\renewcommand{\algorithmicrequire}{\textbf{Input:}}
		\renewcommand{\algorithmicensure}{\textbf{Output:}}
		\caption{H$^2$SGNN for node classification} 
	\label{alg:Framwork} 
		\begin{algorithmic}[1]
			\REQUIRE ~~\\ 
               Raw features of nodes: \textbf{X}, learnable weight matrix: \textbf{W}, order: \textit{K}, number of meta-path: \textit{R}, meta-path matrix: $\mathbf{\hat A}$ and learnable coefficients: $\alpha_{i,k}$, $\beta_i$, $\gamma_k$ where $i\in \{1,...,R\}$ and $k\in \{0,...,K\}$.\\
			\ENSURE ~~\\ 
                Node classification prediction results: $pred$. \\
   
                 \STATE  \% \textit{Local Individual Filtering} \\

        
                \FOR {$i \in \{1,...,R\}$}
                \FOR {$k \in \{0,...,K\}$}
                \STATE $\mathbf{Z}_{i,k}= \alpha_{i,k} h_{i,k}(\mathbf{\hat A}_{i}) \mathbf{X} \mathbf{W}$ 
                \ENDFOR
                \ENDFOR
                \STATE local filtering representation: $\mathbf{Z}_{l}  = \sum_{i=1}^R \sum_{k=0}^K \alpha_{i,k} \mathbf{Z}_{i,k} $ \\
                 \STATE  \% \textit{Global Hybrid Filtering}  \\
                \STATE Constructing the global matrix: $\mathbb{A}=\sum_{i=1}^{R} \beta_i \mathbf{\hat A}_i$
                \FOR {$k \in \{0,...,K\}$}
                \STATE $\mathbf{Z}_{g,k}=\sum_{k=0}^K \gamma_{k} g_k(\mathbb{A}) \mathbf{X} \mathbf{W}$ \\
                \ENDFOR
                \STATE global filtering representation: $\mathbf{Z}_{g}  = \sum_{k=0}^K \mathbf{Z}_{g,k}$ \\
                \STATE final representation: $\mathbf{Z} = \mathbf{Z}_{l} + \mathbf{Z}_{g}$ \\
                \STATE $\hat{\mathbf{y}}=\operatorname{MLP}\left(\mathbf{Z} ; \theta\right)$ \\
                \STATE loss: $\mathcal{L}=\sum_{j \in \mathcal{V}_{\text {train }}} \operatorname{CrossEntropy}\left(\hat{\mathbf{y}}^j, \mathbf{y}^j\right)$ \\
                \STATE \textbf{return} $pred$
            
		\end{algorithmic}
	\end{algorithm}
Algorithm \ref{alg:Framwork} presents our overall training procedure in pseudocode.
First, in the preprocessing step, we obtain different meta-path matrices $\mathbf{\hat A}_i, i\in \{0,...,R\}$ from the heterogeneous graph. Then, for each meta-path matrix $\mathbf{\hat A}_i$, we perform $K$ order filtering and multiply it with the corresponding learnable coefficient $\alpha_{i,k}$ (lines 2-6 in Algorithm 1). Then we add the results of each order of filtering for each meta-path to get the local filtering representation $\mathbf{Z}_{l}$ (line 7 in Algorithm 1). After that, we apply a learnable coefficient $\beta_i$ to each meta-path matrix to construct the global matrix $\mathbb{A}$ (line 9 in Algorithm 1). Then we perform a global filtering on the global matrix (lines 10-12 in Algorithm 1). Finally, we add local and global filtering representation to get the final representation $\mathbf{Z}$, and then perform node classification (lines 13-16 in Algorithm 1).

\section{Detailed Implementation of Polynomial Bases}
The three polynomial bases are defined as follows:

\textbf{Monomial polynomial} represented by GPRGNN \cite{GPR-GNN} directly assigns a learnable coefficient to each order of the normalized adjacency matrix $ \mathbf{\hat A}$, and its filter function is defined as $g_{\gamma, K}(x)=\sum_{k=0}^K \gamma_k x^k$ where $\gamma_k$ are learnable parameters. Thus, when adopting monomial polynomials, the $k$-th basis $h_{i,k}(\cdot)$ of the $i$-th polynomial in Eq. (\ref{eq:local_individual_filtering}) is:

\begin{equation}
h_{i,k}(\mathbf{\hat A}_{i}) = \mathbf{\hat A}_{i}^k.
\end{equation}

\textbf{Jacobi polynomial} \cite{JacobiConv} can adapt to a wide range of weight functions due to its orthogonality and flexibility. The iterative process of Jacobi polynomial can be defined as:
\begin{equation}
\begin{aligned}
& P_0^{a, b}(x)=1 , \\
& P_1^{a, b}(x)=0.5 a-0.5 b+(0.5 a+0.5 b+1) x , \\
& P_k^{a, b}(x)=(2 k+a+b-1) \\
& \cdot \frac{(2 k+a+b)(2 k+a+b-2) x +a^2-b^2}{2 k(k+a+b)(2 k+a+b-2)} P_{k-1}^{a, b}(x) \\
& -\frac{(k+a-1)(k+b-1)(2 k+a+b)}{k(k+a+b)(2 k+a+b-2)} P_{k-2}^{a, b}(x) ,
\end{aligned}
\end{equation}
where $a$ and $b$ are tunable hyperparameters. Thus, when adopting Legendre polynomials, the $k$-th basis $h_{i,k}(\cdot)$ of the polynomial of the $i$-th meta-path in Eq. (\ref{eq:local_individual_filtering}) is:
\begin{equation}
h_{i,k}(\mathbf{\hat A}_{i}) = P_k^{a, b}(\mathbf{\hat A}_{i}).
\end{equation}

\textbf{Legendre polynomial} \cite{Legendre} has a fixed weight function: 1, compared to Jacobi polynomial. The recursive formula for Legendre polynomials is:
\begin{equation}
\begin{aligned}
& P_0(x)= 1 , \\
& P_1(x)= x , \\
& P_{k+1}(x)=\frac{(2 k+1) x P_k(x)-nkP_{k-1}(x)}{(k+1)}.
\end{aligned}
\end{equation}

Then, when adopting Legendre polynomials, the $k$-th basis $h_{i,k}(\cdot)$ of the polynomial of the $i$-th meta-path in Eq. (\ref{eq:local_individual_filtering}) is:
\begin{equation}
h_{i,k}(\mathbf{\hat A}_{i}) = P_k(\mathbf{\hat A}_{i}).
\end{equation}

Note that Legendre polynomials are special cases of Jacobi polynomials when $a=b=0$. However, due to their different weight functions and calculations, their performance may be different.

\section{Experimental Details}
\subsection{Datasets}
The four datasets used in this paper are described as follows:
\begin{itemize}
    \item \textbf{DBLP}~\cite{hgb} is a computer science bibliography website containing papers published between 1994 and 2014 from 20 conferences across four research fields. The dataset consists of four types of nodes: Authors (A), Papers (P), Terms (T), and Venues (V). We choose $APA$, $APTPA$, $APVPA$ as meta-paths. 
    
    \item \textbf{ACM}~\cite{hgb} is an academic citation network that includes papers from three categories: Databases, Wireless Communication, and Data Mining. The dataset contains four types of nodes: Authors (A), Papers (P), Subjects (S), and Fields (F). Pre-defined meta-paths include $PPP$, $PAP$, $PCP$, and $PKP$.
    
    \item \textbf{IMDB}~\cite{hgb} is an online platform that provides information about movies and related details. Movies are categorized into five genres: Action, Comedy, Drama, Romance, and Thriller. The dataset includes four types of nodes: Movies (M), Directors (D), Actors (A), and Keywords (K). The meta-paths we use include $MDM$, $MAM$, and $MKM$.
    
    \item \textbf{AMiner}~\cite{HeCo} is another academic citation network consisting of four types of papers. The dataset includes three types of nodes: Authors (A), Papers (P), and References (R). The meta-paths are $PRP$ and $PAP$.
\end{itemize}

\subsection{Hyperparameters}
The hyperparameter settings of the proposed H$^2$SGNN on four datasets such as order $K$, learning rate, local polynomial, and global polynomial are shown in Table \ref{tb:hyperparameters}. For more details on hyperparameters, please refer to our code.
\begin{table}[ht]
    \centering
    \caption{The hyperparameters of H$^2$SGNN for node classification on four datasets.}
    \resizebox{0.48\textwidth}{!}{
    \begin{tabular}{l c c c c c c c c c}
    \toprule
        Dataset  &local polynomial & global polynomial  &$K$ &learning rate  \\ 
        \midrule
        DBLP   &Legendre&GPRGNN  &6  &0.005\\
        ACM   &Jacobi &GPRGNN &10  &0.0005 \\
        IMDB   &GPRGNN &GPRGNN&10  &0.0005 \\
        AMiner  &GPRGNN &GPRGNN  &10  &0.001\\
        \bottomrule
    \end{tabular}
    }
    \label{tb:hyperparameters}
\end{table}

\subsection{Further Details of Baselines}
We compare our model with the following four types of baselines:

\textit{Homogeneous Homophilic GNNs:} \textbf{GCN} \cite{GCN} uses a simplified first-order Chebyshev polynomial, which is shown to be a low-pass filter. \textbf{GAT}\cite{GAT} uses an attention mechanism to aggregate adjacent nodes. 

\textit{Homogeneous Heterophilic GNNs:} \textbf{GPRGNN} \cite{GPR-GNN} learns a polynomial filter by directly performing gradient descent on the polynomial coefficients.  \textbf{ChebNetII} \cite{ChebNetII} uses Chebyshev interpolation to learn filters.

\textit{Heterogeneous Homophilic GNNs:} \textbf{RGCN}~\cite{RGCN} extends GCN~\cite{GCN} by applying edge type-specific graph convolutions to heterogeneous graphs. \textbf{HAN} \cite{HAN} leverages hierarchical attention to describe node-level and semantic-level structures. \textbf{GTN}~\cite{gtn} employs soft sub-graph selection and matrix multiplication to create neighbor graphs. \textbf{SimpleHGN}~\cite{hgb} incorporates a multi-layer GAT network, utilizing attention based on node features and learnable edge-type embeddings. \textbf{MHGCN}~\cite{mhgcn} learns summation weights directly and uses GCN's convolution for feature aggregation. \textbf{HALO} \cite{halo} proposes a novel heterogeneous GNN architecture in which layers are derived from optimization steps that descend a novel relation-aware energy function.  \textbf{SeHGNN} \cite{sehgnn} utilizes predetermined meta-paths for neighbor aggregations and applies a transformer-based approach. \textbf{MAGNN} \cite{magnn} encodes information from manually selected meta-paths instead of just focusing on endpoints. 

\textit{Heterogeneous Heterophilic GNNs:}
\textbf{HDHGR} \cite{HDHGR} proposes a homophily-oriented deep heterogeneous graph rewiring method. \textbf{Hetero2Net} \cite{Hetero2Net} proposes a heterophily-aware heterogeneous GNN that incorporates both masked meta-path prediction and masked label prediction tasks to handle heterophilic heterogeneous graphs. \textbf{PSHGCN} \cite{PSHGCN} proposes positive non-commutative polynomials to design positive spectral non-commutative graph convolution based on a unified graph optimization framework. This is the most competitive baseline as it claims to have high expressive power.

\section{More Experimental Results}
\subsection{Parameter Sensitivity Analysis on Polynomial Basis}

\begin{figure}[t]   
	\centering 
	\includegraphics[width=\linewidth]{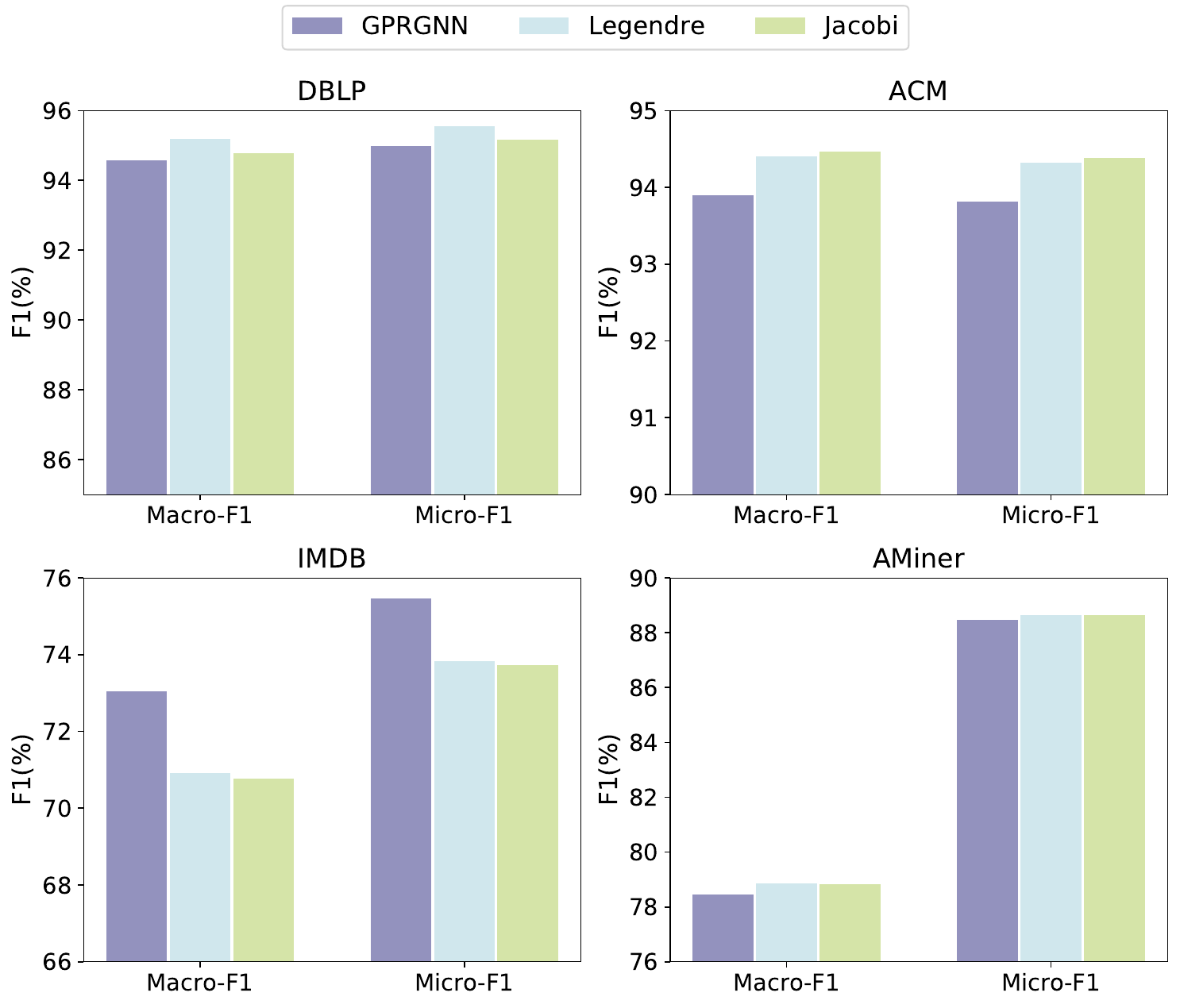}
	\caption{Effect of different polynomials on model performance.}
 \label{fig:poly}
\end{figure}

In Eq. (\ref{eq:local_individual_filtering}) of subsection \ref{local_individual_filtering}, we select different polynomial bases as local independent filters for different datasets. Here we analyze the impact of different polynomial bases such as GPRGNN \cite{GPR-GNN}, Legendre \cite{Legendre} and Jacobi \cite{JacobiConv} on model performance. We use these three polynomials to perform node classification tasks on each dataset, as depicted in Figure \ref{fig:poly}. We can conclude that, except for the IMDB dataset, the performance of local independent filtering using different polynomials on the three datasets is similar. This shows that the expressive power of different polynomials is the same under the same order $K$.  As for the IMDB dataset, since its meta-paths all have low homophily, we speculate that monomial polynomial GPRGNN is more suitable for this dataset and thus outperforms the other two polynomial bases.


\label{sec:abl}

\begin{figure}[t]   
	\centering 
	\includegraphics[width=\linewidth]{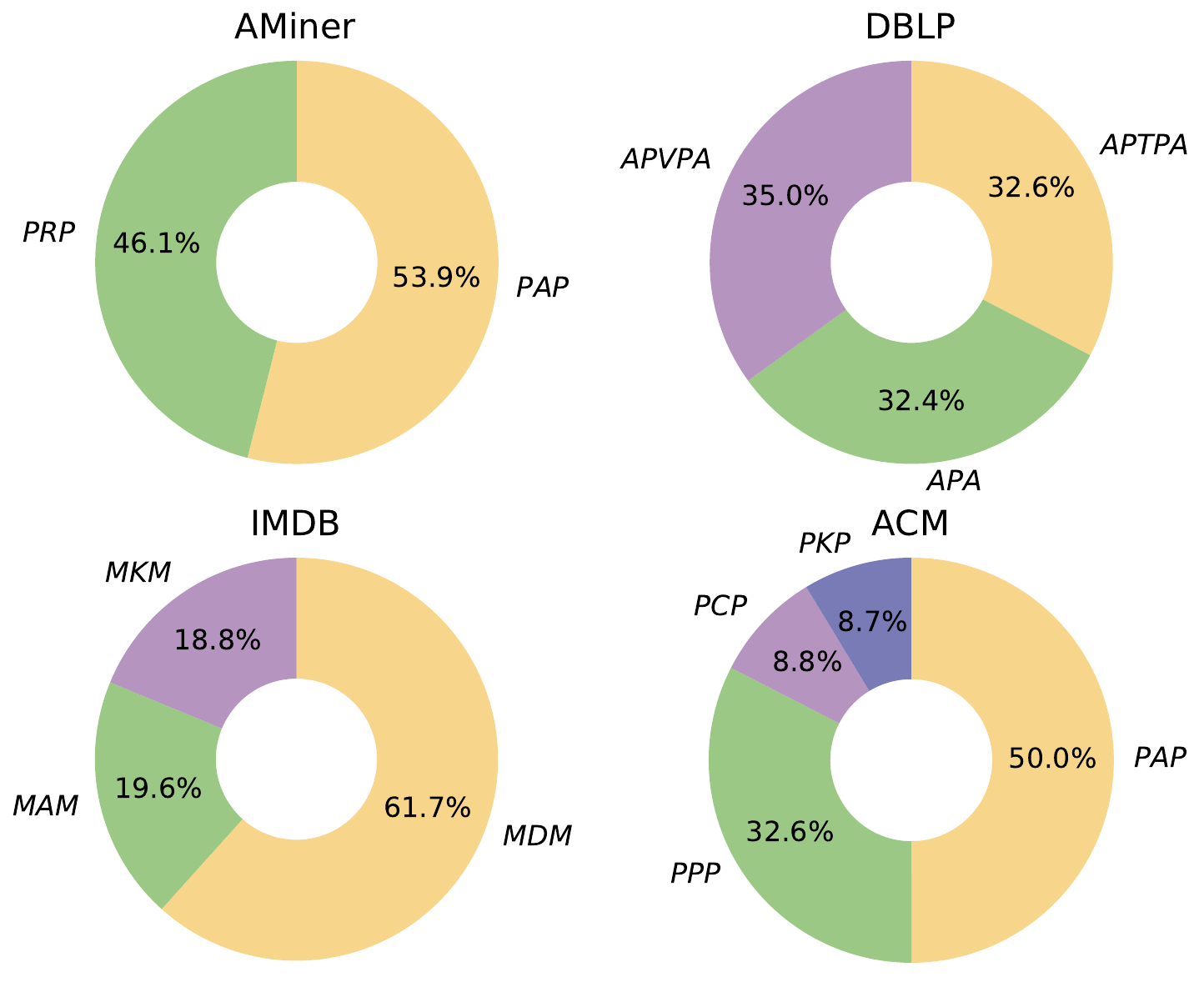}
	\caption{The percentage of $\beta_i$ learned by each meta-path under different datasets.}
 \label{fig:beta}
\end{figure}

\subsection{Importance Analysis of Each Meta-path}
In the \ref{Global_Hybrid_Filtering} Global Hybrid Filter subsection, to measure the importance of each meta-path, we add a learnable parameter to each meta-path to form a global adjacency matrix. As presented in Figure \ref{fig:beta}, we show the percentage of coefficients $\beta_i$ learned for each meta-path. The coefficients learned for each meta-path in the AMiner and DBLP datasets are similar, which shows that each meta-path is almost equally important in these two datasets. However, in the IMDB and ACM datasets, there is only one dominant meta-path, such as the MDM meta-path of the IMDB dataset and the $PAP$ meta-path of the ACM dataset. We find that the homophily of these two meta-paths is relatively high. Therefore, we speculate that meta-paths with higher homophily may be easier to learn, so the MDM meta-path of IMDB and the $PAP$ meta-path of ACM are dominant.

\section{Related Work}
This section introduces related research, including spectral graph neural networks, heterogeneous homophilic GNNs, and heterogeneous heterophilic GNNs.
\subsection{Spectral Graph Neural Network}


According to whether the filter can be learned, the spectral GNNs can be divided into \emph{pre-defined filters} and \emph{learnable filters}. In the category of pre-defined filters, GCN \cite{GCN} uses a simplified first-order Chebyshev polynomial. APPNP \cite{PPNP} utilizes Personalized Page Rank to set the weight of the filter. In the category of learnable filters. ChebNet \cite{Chebyshev} uses Chebyshev polynomials with learnable coefficients. GPR-GNN \cite{GPR-GNN} extends APPNP by directly parameterizing its weights. BernNet \cite{BernNet} uses Bernstein polynomials to learn filters and forces all coefficients positive. JacobiConv \cite{JacobiConv} adopts an orthogonal and flexible Jacobi basis to accommodate a wide range of weight functions. ChebNetII \cite{ChebNetII} uses Chebyshev interpolation to learn filters. Recently, some works have applied spectral GNNs to node-level filtering. For example, DSF \cite{DSF} proposes a novel diversified spectral filtering framework that automatically learns node-specific filter weights. NFGNN \cite{NFGNN} is provided with the capability of precise local node positioning via the generalized translated operator, thus discriminating the variations of local homophily patterns adaptively.





Some work has explored spectral GNNs on multigraphs or heterogeneous graphs. For example, MGNN \cite{MGNN} develops convolutional information processing on multigraphs and introduces convolutional multigraph neural networks. PSHGCN \cite{PSHGCN} proposes positive non-commutative polynomials to design positive spectral non-commutative graph convolution based on a unified graph optimization framework. However, the exponential growth of parameters and memory limits its application. Moreover, the multivariate polynomials are difficult to explain complex and diverse graph filters, such as low-pass, high-pass, band-pass, etc. 

\subsection{Heterogeneous Homophilic GNNs}
According to the way of processing different semantics, heterogeneous graph neural networks can be broadly categorized into meta-path-based and meta-path-free methods. Meta-path-based methods use pre-defined meta-paths to propagate and aggregate neighbor features. For example, HAN \cite{HAN} leverages hierarchical attention to describe node-level and semantic-level structures. MAGNN \cite{magnn} improves HAN by introducing meta-path-based aggregation to learn semantic messages from multiple meta-paths. SeHGNN~\cite{sehgnn} employs pre-defined meta-paths for neighbor aggregation and incorporates a transformer-based method. In addition, some methods \cite{mao2024hetfs,mao2025fhge,mao2025mapn} focus on meta-path search to improve performance

Meta-path-free methods extend message-passing GNNs to heterogeneous graphs without manually designed meta-paths. For example, RGCN~\cite{RGCN} extends GCN~\cite{GCN} by applying edge type-specific graph convolutions to heterogeneous graphs. GTN~\cite{gtn} employs soft sub-graph selection and matrix multiplication to create neighbor graphs. SimpleHGN~\cite{hgb} utilizes attention based on node features and learnable edge-type embeddings. MHGCN~\cite{mhgcn} learns summation weights directly and uses GCN's convolution for feature aggregation. EMRGNN~\cite{emrgnn} and HALO~\cite{halo} propose optimization objectives for heterogeneous graphs and design their architectures by addressing these optimization problems. HINormer~\cite{hinormer} combines a local structure encoder and a relation encoder, using a graph Transformer to learn node embeddings. However, none of these methods consider the heterophily in heterogeneous graphs.

\subsection{Heterogeneous Heterophilic GNNs}
HDHGR \cite{HDHGR} notices the heterophily phenomenon in heterogeneous graphs, and proposes a homophily-oriented deep heterogeneous graph rewiring method. Hetero2Net 
\cite{Hetero2Net} proposes a heterophily-aware heterogeneous GNN that incorporates both masked meta-path prediction and masked label prediction tasks to handle heterophilic heterogeneous graphs. H$^2$GB \cite{H2GB} develops the first graph benchmark that combines heterophily and heterogeneity across various domains, and proposes H$^2$G-former handle heterophilic and heterogeneous graphs effectively. However, the above methods lack expressiveness since they only consider specific meta-paths, thus resulting in suboptimal performance.



\end{document}